\documentclass{article}

\usepackage{microtype}
\usepackage{graphicx}
\usepackage{booktabs}
\usepackage{hyperref}
\usepackage[preprint]{icml2026}
\usepackage{amsmath}
\usepackage{amssymb}
\usepackage{mathtools}
\usepackage{amsthm}
\usepackage{float}
\usepackage{capt-of}
\usepackage{multirow}
\usepackage{makecell}
\usepackage{algorithm}
\usepackage{algorithmic}
\usepackage{xcolor}
\usepackage{colortbl}
\usepackage{tcolorbox}
\usepackage{listings}

\newcommand{\Description}[1]{}

\tcbset{
  promptbox/.style={
    colback=gray!5,
    colframe=black!75,
    fonttitle=\bfseries,
    boxrule=0.5mm,
    arc=2mm
  }
}

\newcommand{\vecv}[1]{\mathbf{#1}}
\newcommand{\mat}[1]{\mathbf{#1}}
\newcommand{\R}{\mathbb{R}}
\newcommand{\ip}[2]{\langle #1, #2 \rangle}
\newcommand{\norm}[1]{\lVert #1 \rVert}
\DeclareMathOperator*{\argmax}{arg\,max}

\icmltitlerunning{VideoDetective}

\makeatletter
\global\icml@noticeprintedtrue
\makeatother

\begin{document}
\twocolumn[
\icmltitle{VideoDetective: Clue Hunting via both Extrinsic Query and Intrinsic Relevance for Long Video Understanding}

\begin{center}
Ruoliu Yang\textsuperscript{1} \quad
Chu Wu\textsuperscript{1} \quad
Caifeng Shan\textsuperscript{1} \quad
Ran He\textsuperscript{2} \quad
Chaoyou Fu\textsuperscript{\textdagger\,1}\\
\textsuperscript{1}Nanjing University\\
\textsuperscript{2}Institute of Automation, Chinese Academy of Sciences\\
\texttt{yangruoliu1@gmail.com}, \texttt{bradyfu24@gmail.com}\\
\url{https://videodetective.github.io/}
\end{center}

\icmlkeywords{Long Video Understanding, Multimodal Large Language Models, Video Question Answering}

\vskip 0.3in
]

\begingroup
\renewcommand{\thefootnote}{\fnsymbol{footnote}}
\footnotetext[2]{Corresponding author.}
\endgroup
\setcounter{footnote}{0}

\begin{abstract}
Long video understanding remains challenging for multimodal large language models (MLLMs) due to limited context windows, which necessitate identifying sparse query-relevant video segments. However, existing methods predominantly localize clues based solely on the query, overlooking the video’s intrinsic structure and varying relevance across segments. To address this, we propose \textbf{VideoDetective}, a framework that integrates query-to-segment relevance and inter-segment affinity for effective clue hunting in long-video understanding. Specifically, we divide a video into various segments and represent them as a visual--temporal affinity graph built from visual similarity and temporal proximity. We then perform a Hypothesis--Verification--Refinement loop to estimate relevance scores of observed segments to the query and propagate them to unseen segments, yielding a global relevance distribution that guides the localization of the most critical segments for final answering with sparse observation. Experiments show our method consistently achieves substantial gains across a wide range of mainstream MLLMs on representative benchmarks, with accuracy improvements of up to 7.5\% on VideoMME-long. Our code is available at \url{https://videodetective.github.io/}.

\end{abstract}

\begin{figure*}[t]
  \centering
  \includegraphics[width=\textwidth]{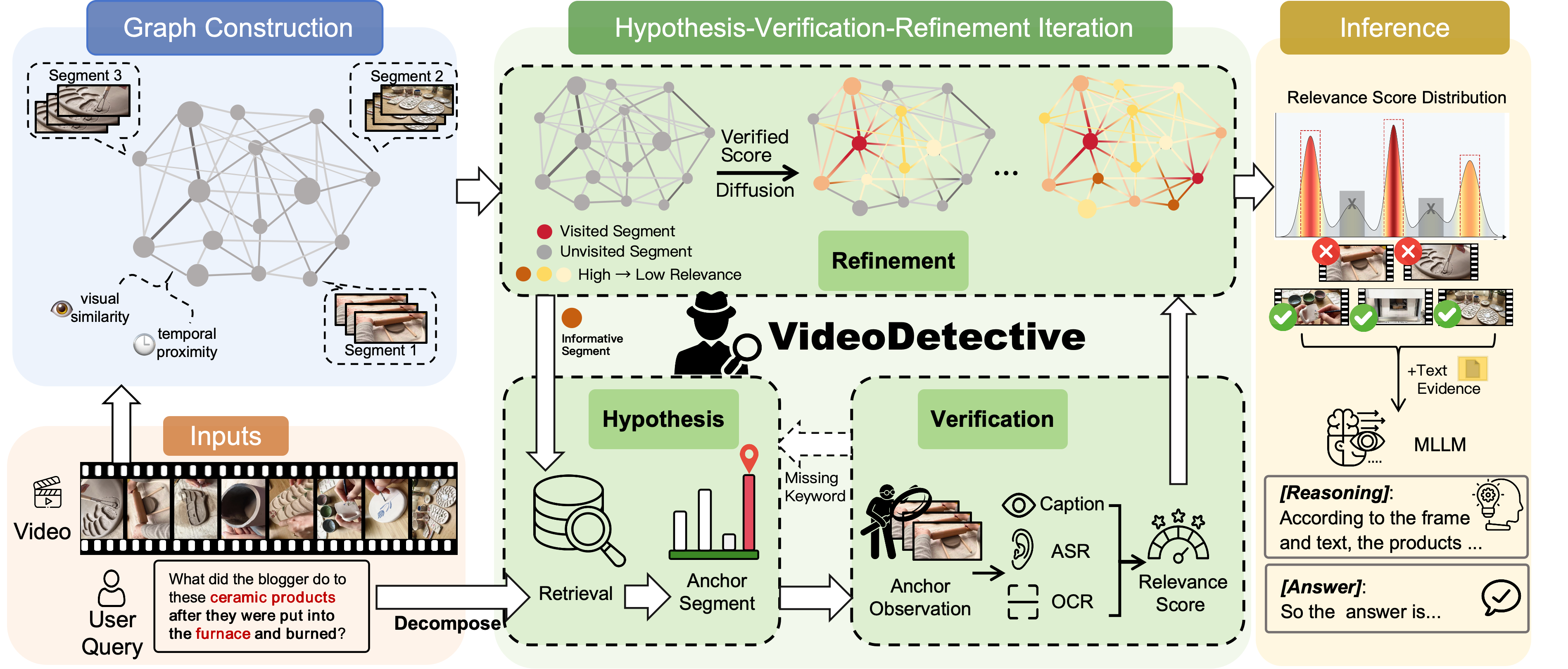}
  \caption{\textbf{Overview of VideoDetective.} Given a query, we (1) divide the video into segments and construct a spatio-temporal affinity graph from visual similarity and temporal proximity; (2) iteratively observe video segments and propagate the relevance scores over the graph to update a global belief field, guiding next observation via a hypothesis–verification–refinement loop to recover missing clues; and (3) aggregate a compact multimodal evidence set (query-relevant frames + related text) for MLLM to produce a clue-grounded answer.}
  \label{fig:overview}
\end{figure*}

\section{Introduction}

Long video understanding has become a central topic in the multimodal community, and a growing number of MLLMs tailored for long-video understanding~\cite{chen2024longvila,zhang2024long,shen2025long,shu2025video} have emerged. Despite this progress, processing massive information within limited context windows remains a critical challenge. As a result, many query-driven approaches focus on locating only the query-relevant clue segments, thereby substantially reducing the effective context length. However, reliably localizing such clues without exhaustively understanding the entire video is inherently difficult, especially for questions requiring complex reasoning.

Most existing methods \cite{wang2025models,liu2025towards} a unidirectional query-to-video search paradigm, matching frames or segments as clues purely based on query information. For example, keyframe selection methods~\cite{park2026too,tang2025adaptive} aim to sample frames with more significant visual information; retrieval-based methods~\cite{luo2024video,jeong2025videorag} convert multimodal video content into text and retrieve clues via textual similarity; and agent approaches~\cite{fan2024videoagent,wang2024videoagent,wang2025active,yuan2025videodeepresearch,zhi2025videoagent2} leverage LLM-based reasoning and external tools to iteratively collect and interpret clues. However, these paradigms share a common limitation: they largely emphasize query-to-content matching while overlooking the video’s intrinsic structures. A video is not merely a linear sequence of isolated frames; it exhibits coherent temporal dynamics and causal continuity. Such internal structure can be exploited to “see the whole from a part,” enabling models to maintain global understanding from sparse observations.

Motivated by this insight, we avoid assuming that a single, prior-driven step can directly pinpoint the truly informative regions, or that the process must restart from scratch once an early guess proves incorrect. Instead, we jointly leverage the query and the video’s intrinsic inter-segment correlations, using sparse observations to model the query-relevance distribution over the entire video. In this way, each observed segment contributes information gain as much as possible under a limited observation budget.

We propose VideoDetective, an inference framework that integrates both extrinsic query relevance and intrinsic video correlations to more accurately localize true clue segments, achieving ``\emph{See Less but Know More}''. Specifically, VideoDetective models the video as a Spatio-Temporal Affinity Graph, explicitly encoding both visual semantics and temporal continuity. Guided by this graph, the framework executes an iterative ``Hypothesis-Verification-Refinement" loop: (1) \textbf{Hypothesis}: initially choose anchor segments based on query-guided prior similarity and iteratively select the next most informative segments as the anchor; (2) \textbf{Verification}: extract multi-source information (e.g., visual captions, OCR, ASR) from anchor segments to verify their local relevance and compute clue scores; (3) \textbf{Refinement}: propagate the relevance of visited segments to unvisited ones via graph diffusion~\cite{zhou2003learning,kipf2017semi} thereby updating the \textit{global belief field} (i.e., a global relevance map over video segments).  In summary:
\begin{itemize}
    \item We propose a long-video inference framework that integrates extrinsic query  with intrinsic video structure. By modeling the video as a Spatio-Temporal Affinity Graph, we exploit internal correlations to guide effective clues localization according to the query.
    
    \item We introduce graph diffusion within a ``Hypothesis-Verification-Refinement'' loop. This mechanism propagates sparse relevance scores from anchor segments across the graph to dynamically update the global belief field, allowing the model to progressively recover global semantic information from sparse observations.
    
    \item We demonstrate that VideoDetective is a plug-and-play framework that consistently improves performance across diverse MLLM backbones. Experiments on representative long-video benchmarks show that our method delivers substantial gains for various baseline models, achieving accuracy improvements of up to 7.5\% on VideoMME-long.
\end{itemize}

\section{Related Work}

\subsection{Multimodal Large Language Models.}
Multimodal Large Language Models (MLLMs)~\cite{hurst2024gpt,lin2024video,bai2025qwen2,comanici2025gemini} combine visual encoders \cite{radford2021learning,zhai2023sigmoid}with LLMs\cite{achiam2023gpt,liu2024deepseek,yang2025qwen3}, achieving remarkable progress in vision-language tasks. However, most MLLMs  struggle with long-form content due to attention complexity and limited context windows. While some recent models~\cite{chen2024longvila,shen2025long,comanici2025gemini} extend context window length to millions of tokens, the computational cost remains prohibitive for dense sampling.

\subsection{Long Video Understanding.}
Long video understanding still remains challenging due to the long temporal horizon and model's limited context budgets. Recent advances in training-free long video understanding methods can be roughly categorized into three main paradigms. \emph{Key-frame sampling and token compression methods}~\cite{park2026too,shen2024longvu,tang2025adaptive,tao2025dycoke,wang2025adaretake} adaptively sample frames or compress tokens to fit context windows, but at the risk of missing critical clues. \emph{Retrieval-augmented methods}~\cite{luo2024video,jeong2025videorag} convert video's content to text and use text-based retrieval to augment generation, but require full-video preprocessing and are limited by information gap from multi-modality to single modality. Recent \emph{agent-based methods}~\cite{fan2024videoagent,wang2024videoagent,wang2025active,yuan2025videodeepresearch,zhi2025videoagent2} explore multi-step reasoning based on LLM planning and tool using, but lack robustness to distractions. 

\label{app:example_figure}


\begin{figure*}[t]
  \centering
  \includegraphics[width=0.9\textwidth]{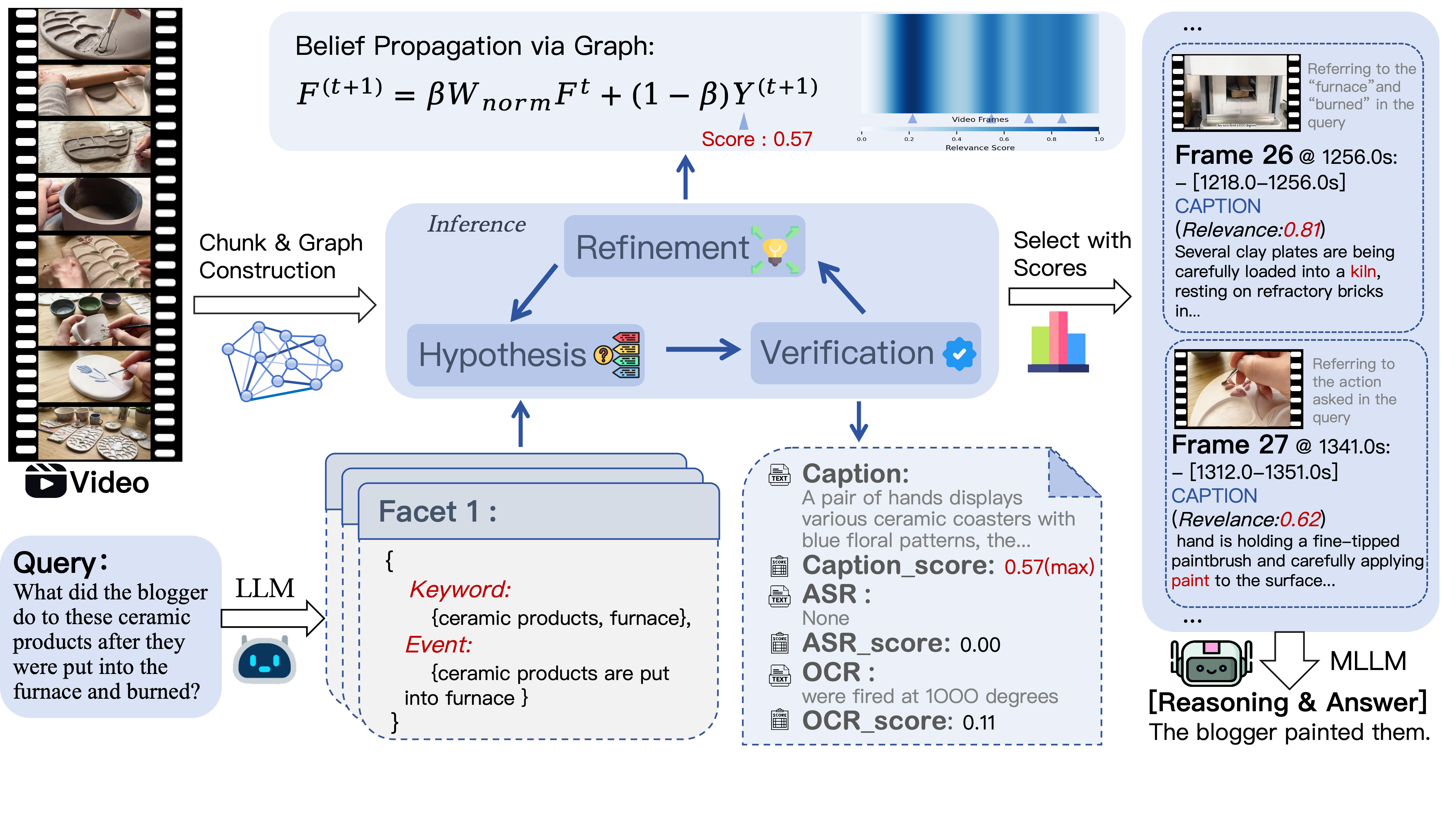}
  \caption{\textbf{A qualitative example of VideoDetective.} It illustrates how VideoDetective processes the input video and query to derive the correct answer. Original video: \url{https://www.youtube.com/watch?v=B6tQyCH5hQM}}
  \label{fig:appendix_example}
\end{figure*}
\section{Methodology}
\label{sec:method}

\subsection{Overview}
\label{sec:overview}

To combine both extrinsic query and intrinsic relevance to localize query-related video segments, we formulate long-video QA as iterative relevance state estimation on a visual--temporal affinity graph $G=(\mathcal{V},\mathcal{E})$ (Algorithm~\ref{alg:videobeacon}).
Given a video $V$, we treat its segments $\{c_i\}_{i=1}^K$ as nodes $\mathcal{V}$ and fuse visual similarity with temporal continuity as edges $\mathcal{E}$.
Two state vectors are maintained at step $t$:
\begin{itemize}
    \item \textbf{Injection Vector} $\vecv{Y}^{(t)} \in \mathbb{R}^K$: 
    A sparse observation vector initialized by priors. It records the verified relevance scores ($Y^{(t)}_i \leftarrow s_i$) at visited segment nodes and serves as the source signal for diffusion.
    \item \textbf{Belief Field} $\vecv{F}^{(t)} \in \mathbb{R}^K$: 
    A dense global relevance scores distribution inferred from $\vecv{Y}^{(t)}$ by propagating information over the affinity graph. Each entry $F^{(t)}_i$ estimates how likely segment $c_i$ contains query-relevant evidence, even if $c_i$ has not been directly observed. 
\end{itemize}
In each iteration, we verify a selected anchor segment (\S\ref{sec:scoring}), update the injection state $\vecv{Y}$, and perform graph diffusion (\S\ref{sec:diffusion}) to refine the belief field $\vecv{F}$.
Finally, we feed top-ranked segments from $\vecv{F}$ into the downstream MLLM for answer generation.

\subsection{Visual-Temporal Affinity Graph Construction}
\label{sec:graph}

To model the continuous global belief field from sparse segment observations, we construct a Visual--Temporal Affinity Graph, which is essentially the topological structure that captures the intrinsic associations between video segments.
This graph defines how relevance scores should propagate from observed anchor segments to unvisited ones.

\subsubsection{Video Segmenting \& Node Representation}
\label{sec:chunk}

To obtain the discrete nodes for our graph, we divide the video into $K$ semantic segments $\{c_i\}_{i=1}^{K}$ based on visual similarity.
Specifically, we extract $T$ frames $\{x_t\}_{t=1}^{T}$ and leverage the SigLIP encoder~\cite{zhai2023sigmoid} to generate frame features $f_t\in\mathbb{R}^D$.
We identify segment boundaries where the cosine similarity between adjacent frames drops below a threshold (i.e., $\langle f_t,f_{t+1}\rangle<\theta_{\mathrm{sim}}$), and subsequently merge fragmented segments shorter than $L_{\min}$.
Finally, each node $i$ is represented by 
$h_i=\operatorname{norm}\left({|c_i|}^{-1}\sum_{t\in c_i} f_t\right)$

\subsubsection{Affinity Matrix}
\label{sec:affinity}

We construct an edge weight matrix $\mat{W}\in\R^{K\times K}$ to define inter-node relations and govern how relevance  diffuse across the graph. The ideal graph structure should satisfy: (1) visually similar segments are highly connected to support cross-temporal information sharing; (2) temporally adjacent segments remain connected to leverage the temporal coherence of events.

\paragraph{Visual affinity}: we define visual affinity as cosine similarity and truncate negative values to avoid spurious anti-correlations, using $\ell_2$-normalized node features $\{h_i\}$:
\begin{equation}
(\mat{W}^{\mathrm{sim}})_{ij}=\max\{0,\ip{h_i}{h_j}\}.
\end{equation}

\paragraph{Temporal affinity}: We model temporal proximity using an exponentially decaying kernel~\cite{belkin2003laplacian}:
\begin{equation}
(\mat{W}^{\mathrm{time}})_{ij}=\exp\!\left(-\frac{|t_i-t_j|}{\tau}\right),
\end{equation}
where $t_i$ denotes the center time of segment $c_i$, and $\tau$ controls the temporal influence range.

\paragraph{Fusion and Sparsification}: 
We synthesize the final affinity graph via a weighted combination $\mat{W}=\alpha \mat{W}^{\mathrm{sim}}+(1-\alpha)\mat{W}^{\mathrm{time}}$, where $\alpha$ balances visual semantics and temporal continuity.
To ensure robust diffusion and mitigate over-smoothing~\cite{li2018deeper}, we remove self-loops ($W_{ii}=0$), sparsify the graph by retaining only the top-$k$ connections per row, and symmetrize the result via $\tilde{\mat{W}}\leftarrow(\tilde{\mat{W}}+\tilde{\mat{W}}^\top)/2$ to enforce bidirectional information flow.

\paragraph{Symmetric normalization}: To ensure diffusion convergence, we adopt the symmetric normalized Laplacian form~\cite{zhou2003learning}. Let $\mat{D}$ be the degree matrix with $D_{ii}=\sum_j \tilde{W}_{ij}$, and define
\begin{equation}
\label{eq:wnorm}
\mat{W}_{\mathrm{norm}} \triangleq \mat{D}^{-\frac{1}{2}}\,\tilde{\mat{W}}\,\mat{D}^{-\frac{1}{2}}.
\end{equation}
This normalization ensures that the spectral radius of $\mat{W}_{\mathrm{norm}}$ is $\le 1$, making the iterative diffusion process converge within bounds~\cite{chung1997spectral}.

\begin{center}
  \begin{minipage}{\linewidth}
  \hrule
  \captionof{algorithm}{Overall pipeline of VideoDetective}
  \label{alg:videobeacon}
  \hrule
  \small
  \begin{algorithmic}[1]
  \REQUIRE Video $V$, Question $q$, Iteration steps budget $B$
  \ENSURE Answer $a$
  
  \STATE \textbf{Preprocessing:}
  \STATE \quad Chunk $V$ into $K$ segments $\{c_i\}_{i=1}^{K}$ with features $\{h_i\}$
  \STATE \quad Generate global event timeline and node descriptions $\{e_i\}$
  \STATE \quad Build affinity graph $\mat{W}$; decompose $q \rightarrow \{(\mathcal{K}_r,\mathcal{P}_r)\}_{r=1}^{R}$
  \STATE \quad Initialize injection scores $\vecv{Y}^{(0)} \leftarrow \textsc{PriorScore}(q,\{e_i\})$; $\vecv{F}^{(0)} \leftarrow \vecv{Y}^{(0)}$
  
  \STATE \textbf{Initialize state:} $\mathcal{M}\gets\{1,\dots,R\}$; $\vecv{v}\gets \vecv{0}$ \COMMENT{$\mathcal{M}$: unresolved facets; $\vecv{v}$: visited mask}
  \STATE \textbf{Initialize anchors:} for each facet $r$, $i^\star \gets \arg\max_{i} (Y^{(0)}_{r})_{i}$
  
  \FOR{$t=1$ to $B$}
      \IF{$\mathcal{M}=\emptyset$ \AND $\sum_{j=1}^{K} v_j = K$}
          \STATE \textbf{break}
      \ENDIF  
      \STATE \textbf{Hypothesis (select next segment):}
      \IF{$\mathcal{M}\neq \emptyset$}
          \STATE $i^\star \gets \arg\max_{j:\, v_j=0,\,\tilde{W}_{ij}>0}\; \tilde{W}_{ij}\cdot F^{(t)}_{j}$
          \COMMENT{next anchor, Eq.~(6)}
          \STATE Select a facet $r\in\mathcal{M}$
  
      \ELSE
          \STATE $\mathcal{M} \gets \mathcal{M}\setminus\{r\}$ \COMMENT{facet verified}
          \STATE $i^\star \gets \arg\max_{j}\; F^{(t-1)}_{j}\cdot(1-v_j)$ \COMMENT{gap filling, Eq.~(7)}
      \ENDIF
  
      \STATE \textbf{Verification (observe and score):}
          \STATE\quad $(s_i,\;need\_more) \gets \textsc{Observe}(i, q, \mathcal{K}_r, \mathcal{P}_r)$
          \COMMENT{extract multimodal evidence and compute score, \S\ref{sec:scoring}}

      \STATE \textbf{Refinement (update hypothesis state):}
           \STATE \quad\textbf{Inject observation:} $Y^{(t)}_i \gets s_i$; $v_i \gets 1$
          \STATE \quad\textbf{Propagate:} $\vecv{F}^{(t)} \gets \textsc{Diffuse}(\vecv{Y}^{(t)}, \mat{W})$

  \ENDFOR
  
  \STATE \textbf{Answer:} $\mathcal{S} \gets \textsc{GraphNMS}(\vecv{F}^{(t)})$;
  \textbf{return} $\textsc{MLLM}(\mathcal{S}, q)$
  \end{algorithmic}
  \hrule
  \end{minipage}
  \end{center}

\subsection{Update Global Belief Field via Hypothesis-Verification-Refinement Iteration}
\label{sec:inference}

Based on the constructed graph, we need to quantify the relevance scores distribution of the entire video with the user query.
To achieve it with sparse observations, we design a \textbf{Hypothesis-Verification-Refinement} loop (Figure~\ref{fig:overview}).
In each iteration, it selects informative anchor segments (Hypothesis), observes the content to verify the presence of query keywords and measure relevance scores (Verification), and propagates these scores across the graph to update the global belief field (Refinement), progressively recovering the complete semantic structure of the video.

\subsubsection{Hypothesis: Prior Injection \& Dynamic Anchor Selection}
The Hypothesis phase is meant for selecting anchor segments that serve as information priors for subsequent verification and refinement.
To ensure precise localizing, we first decompose the user query into semantic facets.
Guided by these facets, we adopt a stage-dependent selection strategy: we employ \textbf{Facet-Guided Initialization} to determine the initial anchor before the iterative loop ($t=0$), and transition to \textbf{Informative Neighbor Exploration} or \textbf{Global Gap Filling} during the iterations ($t>0$).

\noindent\paragraph{Query Decomposition.} \label{sec:facet} 
To ensure precise clues grounding, we employ an LLM to rewrite the query $q$ into $R$ distinct semantic facets $\{f_r\}_{r=1}^R$. For each facet $f_r$, we extract two complementary components: a keyword set $\mathcal{K}_r$ and a semantic description set $\mathcal{P}_r$ :
\begin{equation}
q \xrightarrow{\text{LLM}} \{f_r\}_{r=1}^R, \quad \text{where } f_r = (\mathcal{K}_r, \mathcal{P}_r).
\end{equation}
By isolating these components, we can verify clues for specific entities or events separately, preventing information interference between different segments.

\noindent\paragraph{Selection Policy I: Facet-Guided Initialization.} \label{sec:lexsem} 
To localize initial anchor segment, we compute a hybrid prior score for each facet $r$ by fusing sparse visual matching (keywords to frames) and dense semantic matching (descriptions to timeline)~\cite{arivazhagan2023hybrid}:
\begin{equation}
\label{eq:prior_init}
(Y^{\mathrm{prior}}_r)_i = \alpha \cdot \max_{w\in\mathcal{K}_r} \langle\phi_T(w), h_i\rangle + (1-\alpha) \cdot \max_{p\in\mathcal{P}_r} \langle\psi(p), \psi(e_i)\rangle,
\end{equation}
where $\phi_T$ is the SigLIP text encoder, $\psi$ is the semantic encoder, and $e_i$ are descriptions generated by a coarse VLM scan. We then select the initial anchor to maximize this confidence: $i^{\star(0)}=\operatorname{argmax}_i (Y^{\mathrm{prior}}_r)_i$.

\noindent\paragraph{Selection Policy II: Iterative Active Sampling.}
During the iterative inference process ($t \ge 1$), we dynamically determine the next anchor segment for the following iteration based on the verification feedback from the previous step. We maintain a tracking set $\mathcal{M}$ for unresolved facets.

\textit{Case A: Informative Neighbor Exploration.} 
If the VLM feedback indicates insufficient evidence (e.g., ``missing keywords'') for the current facet $r \in \mathcal{M}$ in ``Verification" stage, we infer that the target event likely resides in the temporal or semantic vicinity of the current anchor. We thus select the next anchor $i^{\star(t)} $from the unvisited neighbors on the affinity graph, prioritizing those with strong connections to the current belief state:
\begin{equation}
\label{eq:iterative_next}
i^{\star(t)} \leftarrow \argmax_{j\in\mathcal{U},\,\tilde{W}_{i^\star j}>0} \left( \tilde{W}_{i^\star j} \cdot F^{(t-1)}_j \right),
\end{equation}
where $\mathcal{U}$ denotes the set of unvisited segments.

\textit{Case B: Global Gap Filling.} 
Conversely, if the evidence for facet $r$ is confirmed, we remove it from $\mathcal{M}$. Once all facets are successfully resolved ($\mathcal{M}=\emptyset$) while the iteration budget remains, we switch to a global exploration strategy to uncover potential blind spots. We greedily select the unvisited node $i^{\star(t)} $ with the highest global belief score:
\begin{equation}
\label{eq:pick_next}
i^{\star(t) }= \argmax_{i} \left( F^{(t-1)}_i \cdot (1-v^{(t-1)}_i) \right),
\end{equation}
where $v^{(t-1)}_i \in \{0,1\}$ is a binary mask indicating whether node $i$ has been visited. This mechanism ensures that promising regions missed by facet-specific searches are eventually captured.

\subsubsection{Verification: Multimodal Evidence Extraction and Relevance Scoring.}
\label{sec:verification}

For each selected anchor node $i$, we perform verification to check whether the observed segment covers the keywords derived from the semantic facet and compute the anchor's relevance score. We extract a multi-source evidence set $\mathcal{E}_i=\{e_i^{\mathrm{cap}},\,e_i^{\mathrm{ocr}},\,e_i^{\mathrm{asr}}\}$:
(1) we employ the VLM to perform a dual-purpose task: generating a detailed scene description while simultaneously verifying alignment with the current facet, explicitly outputting ``\texttt{missing keywords} $x$'' if the keywords $x$ in $\mathcal{K}_r$ are not observed in the visual content;
(2) we extract on-screen text via EasyOCR~\cite{jaidedai_easyocr};
(3) we align pre-generated speech transcripts using Whisper~\cite{radford2023robust}.

\noindent\textbf{Relevance Scoring.} \label{sec:scoring}
Since critical clues are distributed across visual, textual, and acoustic channels, single-modal observations are often insufficient. We extract a multi-source evidence set $E = \{e_{cap}, e_{ocr}, e_{asr}\}$. For each evidence item $e \in E$, we design a ``source-aware'' scoring mechanism to measure its relevance.

\textit{Lexical Similarity.}
We use an IDF-weighted lexical overlap score between evidence text and keywords to calculate lexical similarity:
\begin{equation}
s_{\mathrm{lex}}(e,f_r) = \min\left(1, \frac{\sum_{t\in e \cap \mathcal{K}_r}\mathrm{IDF}(t)}{Z_{\mathrm{lex}}}\right),
\end{equation}
where $Z_{\mathrm{lex}}$ is a normalization constant (see Appendix~\ref{app:similarity}).

\textit{Semantic Similarity.}
We use a text encoder $\psi(\cdot)$ (SigLIP text tower) for dense embeddings and calculate cosine similarity against semantic queries (event descriptions):
\begin{equation}
s_{\mathrm{sem}}(e,f_r) = \max_{p\in\mathcal{P}_r}\frac{\ip{\psi(e)}{\psi(p)}}{\norm{\psi(e)}_2\,\norm{\psi(p)}_2+\epsilon}.
\end{equation}

\textit{Source-aware Fusion.}
Different evidence sources have different signal-to-noise ratios. OCR text is precise but sparse (high precision, low recall) and should trust lexical matching more; visual captions are the opposite (high recall, lower precision) and should trust semantic similarity more. We adopt adaptive weights $\lambda_{src}$ to get the final similarity:
\begin{equation}
s(e,f_r)=\lambda_{\mathrm{src}(e)}\, s_{\mathrm{lex}}(e,f_r) + (1-\lambda_{\mathrm{src}(e)})\, s_{\mathrm{sem}}(e,f_r).
\end{equation}

\textit{Node aggregation.}
For multi-source evidence at node $i$, we take the maximum relevance as their relevance score:
\begin{equation}
\label{eq:node_score}
s_i = \max_{e\in E_i,\, r\in\{1,\dots,R\}} s(e,f_r).
\end{equation}

We then inject the score into the belief field: $\vecv{Y}^{(t+1)}_{i^\star}\leftarrow s_{i^\star}$, mark the node as visited, and propagate via Refinement to update the global belief $\vecv{F}^{(t+1)}$. 
\subsubsection{Refinement: Belief Propagation via Manifold }
\label{sec:diffusion}
We treat the computed relevance score of the observed anchor segment as a injection signal and diffuse it across the affinity graph to infer the relevance scores of other segments.
The resulting global belief field $\vecv{F}$ is optimized to satisfy two properties: (1) \textbf{Consistency} with the sparse observed values in $\vecv{Y}$, and (2) \textbf{Smoothness} with respect to the graph manifold structure.
Formally, we minimize the following cost function~\cite{zhou2003learning,belkin2006manifold}:
\begin{equation}
\label{eq:energy}
\mathcal{J}(\vecv{F}) = \underbrace{\norm{\vecv{F} - \vecv{Y}}_2^2}_{\text{Consistency}} + \mu \underbrace{\vecv{F}^\top \mat{L} \vecv{F}}_{\text{Smoothness on manifold}},
\end{equation}
where $\mathbf{L} = \mathbf{I} - \mathbf{D}^{-1/2} \tilde{\mathbf{W}} \mathbf{D}^{-1/2}$ is the symmetric normalized graph Laplacian. The smoothness term penalizes confidence differences between high-affinity neighbors, enabling relevance to diffuse along visual-temporal paths.

We adopt iterative diffusion for efficiency:
\begin{equation}
\label{eq:diff_iter}
\vecv{F}^{(t+1)} = \beta\,\mat{W}_{\mathrm{norm}}\,\vecv{F}^{(t)} + (1-\beta)\,\vecv{Y}^{(t+1)},
\end{equation}
where $\beta=\mu/(1+\mu)\in(0,1)$ balances smoothness and consistency. With top-$k$ sparsification, $\mat{W}_{\mathrm{norm}}$ has $O(Kk)$ non-zeros; using sparse observation, each iteration costs $O(Kk)$, yielding $O(TKk)$ overall (with $k \ll K$)\cite{yedidia2003understanding}. A detailed derivation of the complexity is deferred to Appendix.

\subsection{Segment Selection via Graph-NMS}
\label{sec:selection}

Upon the completion of the iteration, we obtain the converged global belief field, which serves as the final relevance scores distribution for sampling.
To extract a diverse and representative set of key segments, we apply Graph-NMS~\cite{bodla2017soft}.
This mechanism prioritizes high-confidence regions while enforcing diversity through neighbor suppression on the affinity graph.
Crucially, we explicitly retain the maximum-belief node for each query facet to guarantee that all semantic aspects are covered before feeding the aggregated evidence to the downstream MLLM.

\section{Experiments}

\begin{figure*}[!t]
  \centering
  \includegraphics[width=0.85\textwidth]{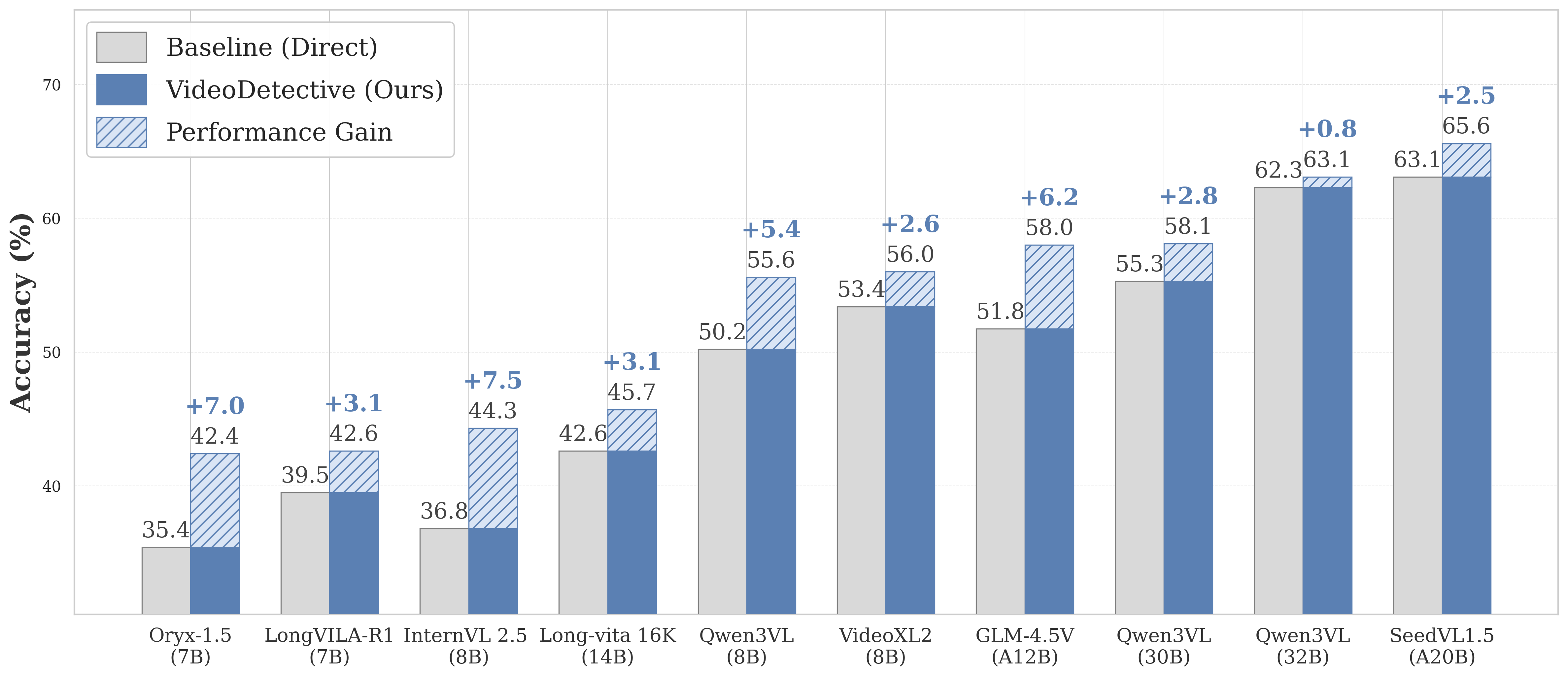}
  \caption{\textbf{Performance improvements across different backbones on VideoMME-long w/o subtitle.} VideoDetective consistently enhances various MLLM across different architectures and parameter scales, demonstrating its plug-and-play capability. For this backbone comparison, VideoXL2 and Oryx-1.5 use 16 sampled frames, InternVL-2.5 uses 8 sampled frames, and all other models use 32 sampled frames; GLM, SeedVL, and Qwen3-VL (30B/32B variants) use Qwen3-30B as the LLM planner, while all other models use Qwen3-8B.}
  \label{fig:backbone_comparison}
\end{figure*}

\subsection{Experiments Setup}

\noindent\paragraph{Benchmarks.} To comprehensively evaluate the overall performance of VideoDetective in long-video understanding, we conduct experiments on four representative benchmarks. Specifically, we evaluate on the long-video subset without subtitles (long subset w/o subtitles) of VideoMME~\cite{fu2025video} and LVBench~\cite{wang2025lvbench} without auxiliary transcripts, and complete evaluations on the validation split (Val split) of LongVideoBench~\cite{wu2024longvideobench} and the test split of MLVU~\cite{zhou2025mlvu}.

\noindent\paragraph{Baselines.} We compare with baselines across three tiers: proprietary models (Gemini-2.5-pro~\cite{comanici2025gemini}, GPT-4o~\cite{hurst2024gpt}, Gemini-1.5-Pro~\cite{team2024gemini}, SeedVL-1.5~\cite{guo2025seed1}), large-scale open-source models ($\ge$72B parameters: Qwen2.5-VL-72B~\cite{bai2025qwen2}, LLaVA-Video-72B~\cite{zhang2024video}), and lightweight open-source models ($<$30B: LongVITA-16k ~\cite{shen2025long}, LongVILA~\cite{chen2024longvila}, InternVL-2.5~\cite{chen2024expanding}, etc.\cite{fu2025vita,li2024llava,shu2025video,zhang2024video,bai2025qwen2,bai2025qwen3vltechnicalreport}). We also apply VideoDetective framework to various backbones (Figure~\ref{fig:backbone_comparison}) to prove its effectiveness  and reproduce other three representative methods with the same backbones for fair comparison.

\noindent\paragraph{Parameters setting.} We set the default inference budget to 10 iterations. In each verification step, the VLM observes a local window of 9 frames with 1fps sampling. For graph construction, we use a sparsity of top-$k=8$ and a temporal decay factor $\tau=30.0$. 

\noindent\paragraph{Evaluation Environment.} API-based models (Gemini~\cite{comanici2025gemini}, Qwen~\cite{bai2025qwen2,bai2025qwen3vltechnicalreport,yang2025qwen3}, SeedVL~\cite{guo2025seed1}, GLM~\cite{hong2025glm} series) are tested via official APIs. Other open-source MLLM backbones are deployed and evaluated on NVIDIA RTX 4090 GPU clusters. 

\subsection{Main Results}
\subsubsection{Generalization across Different Backbones}

To verify the universality of our approach, we applied VideoDetective to a diverse set of MLLM~\cite{chen2024expanding,liu2024oryx,shen2025long,bai2025qwen3vltechnicalreport,qin2025video,hong2025glm,guo2025seed1,chen2025scaling} backbones ranging from 8B to 32B parameters. As illustrated in Figure~\ref{fig:backbone_comparison}, VideoDetective brings a substantial 7.5\% improvement to InternVL-2.5 (8B), 7.0\% to Oryx-1.5 (7B) and robust gains on other baseline models. These results demonstrate that VideoDetective functions as a plug-and-play inference framework that improves long-video performance by jointly leveraging extrinsic query-guided priors and intrinsic manifold propagation. 

\begin{table}[t]
  \centering
  \small
  \setlength{\tabcolsep}{5pt}
  \setlength{\abovecaptionskip}{4pt}
  \setlength{\belowcaptionskip}{0pt}
  \caption{\textbf{Comparison with other baseline methods.} We compare VideoDetective with four representative long-video understanding frameworks using two different backbones (all with 32 frames sampling to answer) on \textbf{VideoMME-long} w/o subtitle. Ours achieves the best performance.}
  \label{tab:framework_comparison}
  \begin{tabular}{lcc}
  \toprule
  \textbf{Backbone (LLM + VLM)} & \textbf{Method} & \textbf{Accuracy (\%)} \\
  \midrule
  \multirow{5}{*}{\makecell[l]{Qwen3-8B\\+ Qwen3VL-8B}} 
  & LVNet & 40.4 \\
  & DVD & 42.6 \\
  & VideoAgent & 42.0 \\
  & VideoRAG & 50.3 \\
  \cmidrule(l){2-3}
  \rowcolor{gray!10} & \textbf{VideoDetective} & \textbf{55.6} \\
  \midrule
  \multirow{5}{*}{\makecell[l]{Qwen3-30B\\+ SeedVL-1.5}} 
  & LVNet & 51.7 \\
  & DVD & 45.4 \\
  & VideoAgent & 51.7 \\
  & VideoRAG & 62.0 \\
  \cmidrule(l){2-3}
  \rowcolor{gray!10} & \textbf{VideoDetective} & \textbf{65.6} \\
  \bottomrule
  \end{tabular}
\end{table}

\subsubsection{Controlled Comparison with Representative Methods}

To validate the independent effectiveness of our framework rather than backbone models, we conduct a fair comparison between VideoDetective and other four representative long-video understanding paradigms—LVNet~\cite{park2026too}, Deep Video Discovery (DVD)~\cite{zhang2025deep}, VideoAgent~\cite{fan2024videoagent}, and VideoRAG~\cite{luo2024video}—all of them unify MLLM and LLM: Qwen3VL-8B and SeedVL-1.5, sampling 32 frames for the final MLLM answer generation across all methods. The experimental results demonstrate that regardless of the strength of the base model, VideoDetective also can unleash its long-video understanding potential and consistently outperforms these representative frameworks across the same backbones.

\subsubsection{Comparison with State-of-the-Art Models}

\begin{table*}[t]
  \centering
  \footnotesize
  \setlength{\tabcolsep}{3pt}
  \caption{\textbf{Comparison with State-of-the-Art Models.} We report the accuracy (\%) on four challenging long-video benchmarks of our methods and other baseline models. And the number of frames \textbf{finally fed to MLLM} to generate answer is 32.}
  \label{tab:main_results}
  \begin{tabular}{lcccccc}
  \toprule
  \multirow{2}{*}{\textbf{Model}} & \multirow{2}{*}{\textbf{Param}} & \multirow{2}{*}{\textbf{Frames}} & \textbf{VideoMME} & \textbf{LVBench} & \textbf{MLVU} & \textbf{LongVideoBench} \\
  & & & \scriptsize{(Long w/o sub)} & \scriptsize{} & \scriptsize{(Test)} & \scriptsize{(Val)} \\
  \midrule
  \multicolumn{7}{l}{\textit{\textbf{Proprietary Models}}} \\
  \midrule
  Gemini-2.5-Pro & - & 32 & 68.7 & 50.1 & 62.2 & 66.8 \\
  GPT-4o & - & 384 & 65.3 & 48.9 & 54.9 & 66.7 \\
  Gemini-1.5-Pro & - & 256 & 67.4 & 33.1 & 53.8 & 64.0 \\
  SeedVL-1.5 & 20B(A) & 32 & 63.1 & 46.1 & 54.9 & 63.8 \\
  \midrule
  \multicolumn{7}{l}{\textit{\textbf{Open-Source Models ($<$ 30B)}}} \\
  \midrule
  LongVITA-16k & 14B & 64 & 54.7 & - & - & 59.4 \\
  LongVILA & 7B & 1fps & 53.0 & - & - & 57.1 \\
  LLaVA-OneVision & 7B & - & 46.7 & - & 47.2 & 56.4 \\
  LLaVA-Video & 7B & 512 & 52.9 & 43.1 & - & 58.2 \\
  VideoXL & 7B & 1fps & 52.3 & 42.9 & 45.5 & 50.7 \\
  Qwen2.5-VL & 7B & 128 & 53.9 & 36.9 & 45.5 & 51.0 \\
  Qwen3-VL & 8B & 32 & 50.2 & 41.1 & 50.1 & 58.9 \\
  InternVL-2.5 & 8B & 32 & 50.8 & 39.9 & 52.8 & 59.2 \\
  VITA-1.5 & 7B & 16 & 47.1 & 37.1 & 39.4 & 53.6 \\
  \rowcolor{gray!10} \textbf{VideoDetective (Qwen3-VL)} & 8B & 32 & \textbf{55.6} & \textbf{43.2} & \textbf{56.3} & \textbf{60.2} \\
  \midrule
  \multicolumn{7}{l}{\textit{\textbf{Open-Source Models ($\ge$ 30B)}}} \\
  \midrule
  Qwen2.5-VL & 72B & 128 & 64.6 & 47.4 & 53.8 & - \\
  LLaVA-Video & 72B & 64 & \textbf{70.3} & 46.1 & - & 63.9 \\
  \rowcolor{gray!10} \textbf{VideoDetective (SeedVL-1.5)} & 20B(A) & 32 & 65.6 & \textbf{51.3} & \textbf{63.8} & \textbf{67.9} \\
  \bottomrule
  \end{tabular}
\end{table*}

As shown in Table~\ref{tab:main_results}, VideoDetective establishes a new state-of-the-art across different parameter scales, both lightweight and the heavier. When equipped with SeedVL-1.5 (20B), our framework achieves 67.9\% accuracy on the challenging LongVideoBench (Val). This performance not only surpasses the significantly larger LLaVA-Video-72B (63.9\%) by a clear margin but also outperforms leading proprietary models such as Gemini-2.5-pro (66.8\%), GPT-4o (66.7\%) and Gemini-1.5-Pro (64.0\%), demonstrating the effectiveness of our framework.

\subsection{Ablation Studies}

\subsubsection{Component Analysis}

\begin{table}[t]
  \centering
  \small
  \setlength{\tabcolsep}{8pt}
  \caption{\textbf{Ablation Study on VideoMME-long w/o subtitle.} Contribution of each core component in VideoDetective.}
  \label{tab:ablation}
  \begin{tabular}{lcc}
  \toprule
  \textbf{Configuration} & \textbf{Accuracy (\%)} & \textbf{$\Delta$} \\
  \midrule
  \rowcolor{gray!10} \textbf{VideoDetective (Full)} & \textbf{55.6} & - \\
  \midrule
  \multicolumn{3}{l}{\textit{\textbf{1. Graph \& Propagation}}} \\
  w/o Graph Propagation & 51.4 & \textcolor{red}{-4.2} \\
  \midrule
  \multicolumn{3}{l}{\textit{\textbf{2. Active Inference}}} \\
  \makecell[l]{w/o Facet Decomposition \&\\ Iterative Refinement} & 47.8 & \textcolor{red}{-7.8} \\
  w/o Iterative Refinement & 51.0 & \textcolor{red}{-4.6} \\
  \midrule
  \multicolumn{3}{l}{\textit{\textbf{3. Multimodal Evidence}}} \\
  w/o Textual Evidence & 49.9 & \textcolor{red}{-5.7} \\
  w/o Optimized Sampling & 50.7 & \textcolor{red}{-4.9} \\
  \midrule
  \textit{Baseline (Direct Inference)} & 50.2 & \textcolor{red}{-5.4} \\
  \bottomrule
  \end{tabular}
\end{table}

To verify the necessity of each core component in VideoDetective, we conduct detailed ablation experiments on the VideoMME-long benchmark without subtitle (Table~\ref{tab:ablation}). We choose the Qwen3VL-8B-Instruct as the multimodal backbone and Qwen3-8B as LLM. For the baseline, we uniformly sample 32 frames as input to Qwen3VL-8B-Instruct.

\paragraph{Impact of Graph Manifold Structure.} Removing the graph propagation mechanism (w/o Propagation) degrades performance by 4.2\%. This confirms that isolated anchor nodes observations are insufficient, and the manifold smoothness constraint is essential for inferring the relevance of unvisited regions based on sparse signals.

\paragraph{Necessity of Facet Decomposition.} Retaining propagation but removing query semantic decomposition (w/o Facet Decomposition) causes accuracy to degrade to 47.8\%, performing even worse than the baseline. This indicates that blind similarity propagation introduces substantial noise. Our semantic facet decomposition acts as a crucial ``compass," ensuring that relevance signals propagate along semantically valid paths rather than visual similarities alone.

\paragraph{Efficiency of Iterative Loop.} The ``hypothesis-verification-refinement" loop is indispensable; replacing it with a single-round observation for each facet(w/o Iterative Refinement) leads to a 4.6\% drop. This validates that our evidence-driven mechanism can effectively correct biases from initial retrieval through iteration.

\paragraph{Complementarity of Multimodal Evidence.} Neither relying solely on visual frames (Visual Only, 49.9\%) nor adding textual evidence (detailed caption + OCR + ASR) which keep the same format as our framework to uniform frame sampling (Both frames and texts, 50.7\%) can achieve optimal performance, verifying the strong complementarity between textual evidence and visual features.

\subsubsection{Modality Scaling Analysis}

\begin{table}[t]
  \centering
  \small
  \setlength{\tabcolsep}{5pt}
  \caption{\textbf{Modality Scaling Analysis.} Performance bottleneck investigation by independently scaling LLM and Visual Encoder.}
  \label{tab:modality_scaling}
  \begin{tabular}{llcc}
  \toprule
  \textbf{LLM} & \textbf{VLM} & \textbf{Acc. (\%)} & \textbf{Gain} \\
  \midrule
  \multicolumn{4}{l}{\textit{Baseline Configuration}} \\
  \quad Qwen3-8B & Qwen3-VL-8B & 55.6 & - \\
  \midrule
  \multicolumn{4}{l}{\textit{Scaling LLM}} \\
  \quad \textbf{Qwen3-30B} & Qwen3-VL-8B & 55.8 & \textcolor{gray}{+0.2} \\
  \midrule
  \multicolumn{4}{l}{\textit{Scaling VLM}} \\
  \quad Qwen3-8B & \textbf{SeedVL-1.5} & \textbf{65.1} & \textbf{\textcolor{teal}{+9.5}} \\
  \midrule
  \multicolumn{4}{l}{\textit{Scaling Both}} \\
  \quad Qwen3-30B & SeedVL-1.5 & 65.6 & \textbf{\textcolor{teal}{+10.0}} \\
  \bottomrule
  \end{tabular}
\end{table}

Finally, we investigate the contribution weights of visual perception and language reasoning to long-video understanding performance (Table~\ref{tab:modality_scaling}). We adopt a strategy of independently scaling the capabilities of the LLM and VLM.

The experimental results reveal asymmetry: when we fix the VLM to 8B and only upgrade the LLM from 8B to 30B, performance almost stagnates (from 55.6\% increasing only marginally to 55.8\%), indicating that an 8B-level LLM already owns sufficient capability to decompose queries. In contrast, when we fix the LLM at lightweight 8B and only upgrade the VLM to the stronger SeedVL-1.5, accuracy achieves a qualitative leap (surging from 55.6\% to 65.1\%, $\Delta$+9.5\%). This powerfully demonstrates that under the VideoDetective framework, the performance ceiling bottleneck still lies in the visual model.

\subsection{Efficiency Analysis}

\begin{figure}[htbp]
  \centering
  \includegraphics[width=0.48\textwidth]{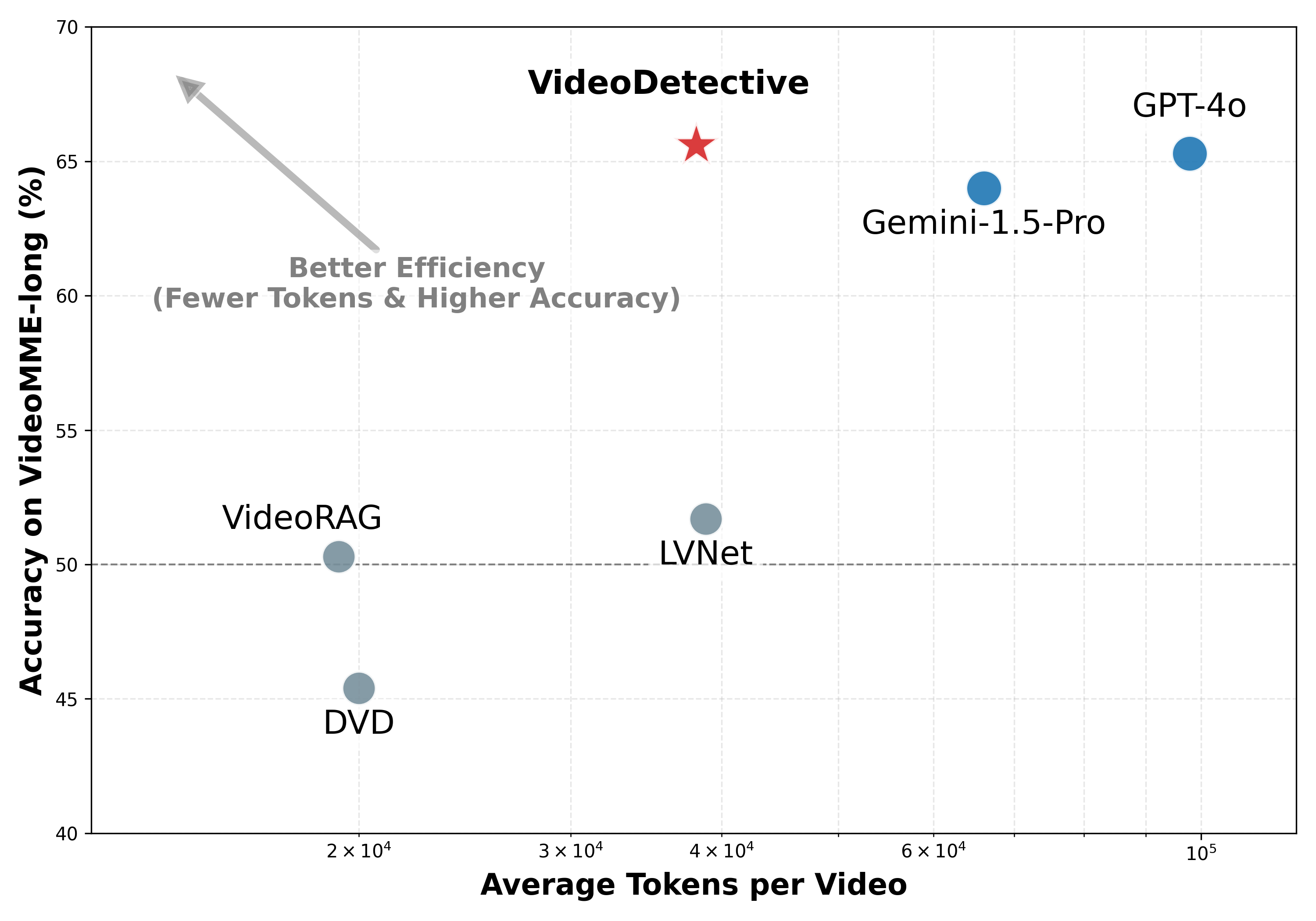}
  \caption{\textbf{Token Efficiency. Comparison of accuracy versus average token consumption(whole pipeline).} VideoDetective achieves the optimal position on the Pareto frontier.}
  \label{fig:efficiency}
\end{figure}

\begin{figure}[t]
  \centering
  \includegraphics[width=\linewidth]{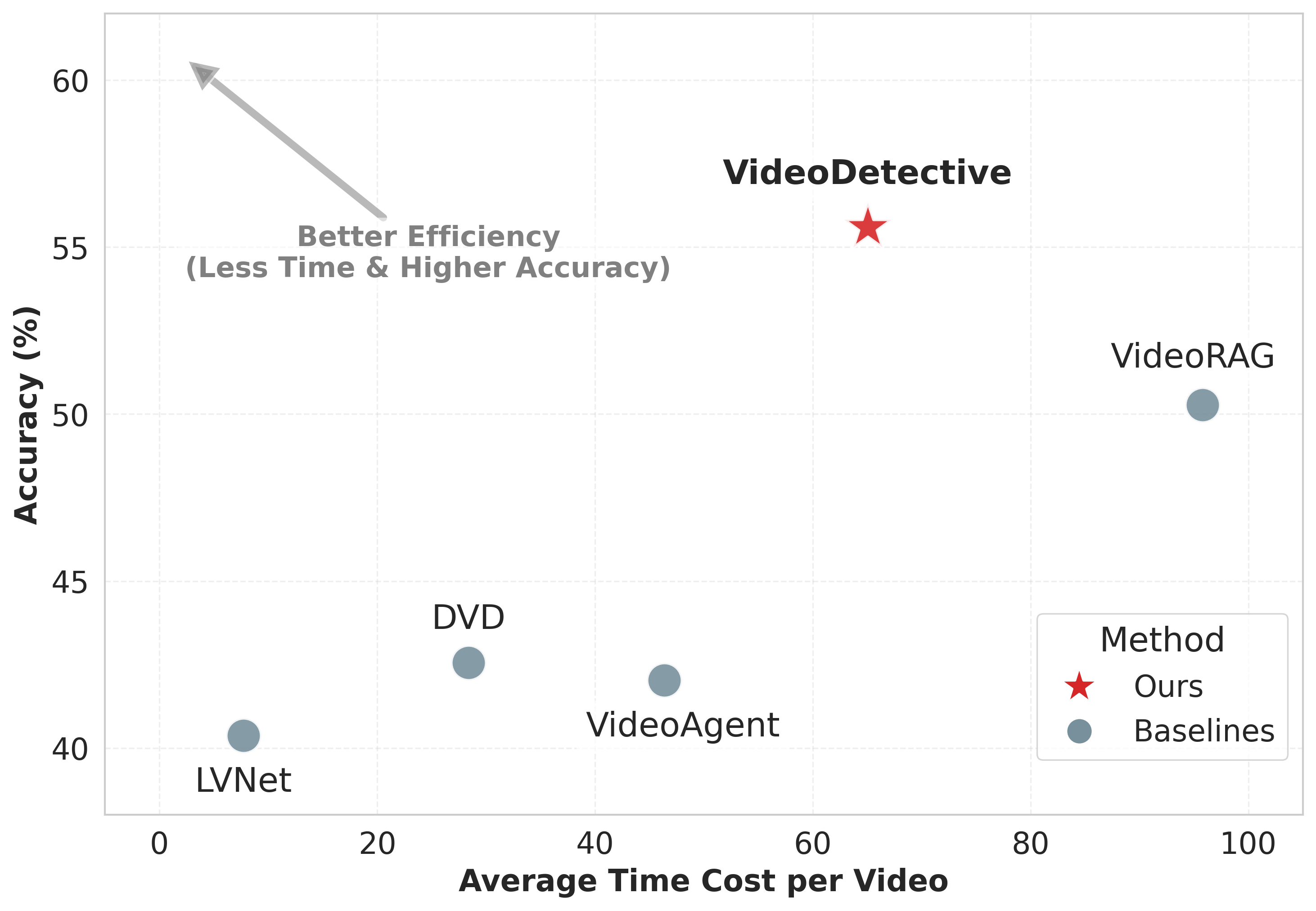}
  \caption{\textbf{Runtime-Accuracy Comparison.} Comparison of average end-to-end complete processing time per video and the corresponding accuracy of representative long-video understanding methods on VideoMME-long w/o subtitle.}
  \label{fig:time_accuracy}
\end{figure}

A comprehensive evaluation of long-video understanding methods requires considering both the computational budget (token consumption) and the actual end-to-end processing time. As illustrated in Figures~\ref{fig:efficiency} and \ref{fig:time_accuracy}, we report the average token consumption and the complete processing time per video on the VideoMME-long benchmark.

\subsubsection{Token Efficiency.} As shown in Figure~\ref{fig:efficiency}, VideoDetective attains competitive accuracy (65.6\%) with a moderate token consumption of $\sim$38k per video. This provides a clear advantage over dense sampling on other open-source and proprietary models like GPT-4o, which require more input tokens ($\sim$10$^5$) to achieve similar performance. Furthermore, compared to other token-efficient methods (e.g., VideoRAG, DVD) with a similar token budget of $\sim$20k, our framework delivers an absolute accuracy gain of over 13\%.

\subsubsection{Runtime-Accuracy Trade-off.} Figure~\ref{fig:time_accuracy} evaluates the complete time, which includes all pipeline stages: offline preprocessing (sampling, graph construction, OCR/ASR), the iterative VLM observation loop, and final answer generation. While straightforward single-step or pure retrieval methods (e.g., LVNet, DVD) exhibit lower latency ($<30$s), their accuracy is limited ($<43\%$). Conversely, heavy retrieval-augmented pipelines like VideoRAG demand nearly 100 seconds per video with sub-optimal results. VideoDetective strikes a better balance: by taking approximately 65 seconds per video to execute the Hypothesis-Verification-Refinement loop and graph diffusion, it achieves a 55.6\% accuracy (under the 8B setting).

\begin{table}[H]
  \centering
  \small
  \setlength{\tabcolsep}{8pt}
  \caption{\textbf{Iteration Budget Analysis.} Effect of different active inference budgets on accuracy and average end-to-end processing time per video on VideoMME-long w/o subtitle.}
  \label{tab:iteration_budget}
  \begin{tabular}{ccc}
  \toprule
  \textbf{Iterations} & \textbf{Accuracy (\%)} & \textbf{Avg. Time / Video} \\
  \midrule
  4 & 53.4 & 38.1 \\
  6 & 53.4 & 47.1 \\
  8 & 52.1 & 56.3 \\
  \rowcolor{gray!10} \textbf{10 (default)} & \textbf{55.6} & \textbf{65.1} \\
  12 & 56.1 & 74.9 \\
  \bottomrule
  \end{tabular}
\end{table}

\begin{table}[H]
  \centering
  \scriptsize
  \setlength{\tabcolsep}{4pt}
  \caption{\textbf{Stage-wise Runtime and Token Breakdown.} Average time and token consumption per video on VideoMME-long w/o subtitle, decomposed into preprocessing, the HVR loop, and final answer generation.}
  \label{tab:stage_breakdown}
  
  \begin{tabular}{@{}lcccc@{}}
  \toprule
  \textbf{Stage} & \textbf{Time} & \makecell{\textbf{Time}\\\textbf{Ratio}} & \textbf{Tokens} & \makecell{\textbf{Tokens}\\\textbf{Ratio}} \\
  \midrule
  \rowcolor{gray!10} \textbf{Total} & \textbf{65.1} & \textbf{100\%}& \textbf{38103} & \textbf{100\%} \\
  Preprocessing & 24.1 & 37\% & 9145 & 24\% \\
  HVR loop & 31.2 & 48\% & 19814 & 52\% \\
  Final answer & 9.8 & 15\% & 9145 & 24\%\\
  \bottomrule
  \end{tabular}
\end{table}

\subsubsection{Iteration Budget Analysis} To further study the trade-off between inference depth and efficiency, we vary the active inference budget from the default 10 iterations to 4, 6, 8, and 12 while keeping all other settings fixed, and report the accuracy and average processing time per video in Table~\ref{tab:iteration_budget}. The default parameter settings have achieved a good balance between accuracy and time cost.

\subsubsection{Stage-wise Cost Breakdown} To better understand how the efficiency budget is distributed, we further decompose the complete pipeline into preprocessing, the Hypothesis-Verification-Refinement (HVR) loop, and final answering, and report both the average time and the average token consumption per video in Table~\ref{tab:stage_breakdown}.

\section{Conclusion}

We present \textsc{VideoDetective}, an inference framework that integrates both extrinsic query relevance and intrinsic video correlations. By modeling a long video as a visual--temporal affinity graph and performing a hypothesis--verification--refinement inference loop, we propagate query-relevance signals from sparse local observations to the entire video, thereby locating critical clues for long-video understanding. Extensive experiments on four challenging benchmarks demonstrate that our approach achieves competitive performance against strong MLLMs and consistently outperforms existing baselines, while maintaining computational efficiency through sparse sampling.

\section{Limitation} Our method relies on the self-reflection capability of VLMs to provide feedback signals (e.g., ``missing keywords''); future work may explore more sophisticated relevance assessment mechanisms for improved robustness.


\bibliographystyle{icml2026}
\bibliography{arxiv_v2}

@article{chen2024longvila,
  title   = {Longvila: Scaling long-context visual language models for long videos},
  author  = {Chen, Yukang and Xue, Fuzhao and Li, Dacheng and Hu, Qinghao and Zhu, Ligeng and Li, Xiuyu and Fang, Yunhao and Tang, Haotian and Yang, Shang and Liu, Zhijian and others},
  journal = {arXiv preprint arXiv:2408.10188},
  year    = {2024}
}

@article{shen2025long,
  title   = {Long-vita: Scaling large multi-modal models to 1 million tokens with leading short-context accuracy},
  author  = {Shen, Yunhang and Fu, Chaoyou and Dong, Shaoqi and Wang, Xiong and Zhang, Yi-Fan and Chen, Peixian and Zhang, Mengdan and Cao, Haoyu and Li, Ke and Lin, Shaohui and others},
  journal = {arXiv preprint arXiv:2502.05177},
  year    = {2025}
}

@inproceedings{shu2025video,
  title     = {Video-xl: Extra-long vision language model for hour-scale video understanding},
  author    = {Shu, Yan and Liu, Zheng and Zhang, Peitian and Qin, Minghao and Zhou, Junjie and Liang, Zhengyang and Huang, Tiejun and Zhao, Bo},
  booktitle = {CVPR},
  year      = {2025}
}

@article{qin2025video,
  title   = {Video-XL-2: Towards Very Long-Video Understanding Through Task-Aware KV Sparsification},
  author  = {Qin, Minghao and Liu, Xiangrui and Liang, Zhengyang and Shu, Yan and Yuan, Huaying and Zhou, Juenjie and Xiao, Shitao and Zhao, Bo and Liu, Zheng},
  journal = {arXiv preprint arXiv:2506.19225},
  year    = {2025}
}

@article{hong2025glm,
  title   = {GLM-4.1 V-Thinking: Towards Versatile Multimodal Reasoning with Scalable Reinforcement Learning},
  author  = {Hong, Wenyi and Yu, Wenmeng and Gu, Xiaotao and Wang, Guo and Gan, Guobing and Tang, Haomiao and Cheng, Jiale and Qi, Ji and Ji, Junhui and Pan, Lihang and others},
  journal = {arXiv preprint arXiv:2507.01006},
  year    = {2025}
}

@article{zhang2024long,
  title   = {Long context transfer from language to vision},
  author  = {Zhang, Peiyuan and Zhang, Kaichen and Li, Bo and Zeng, Guangtao and Yang, Jingkang and Zhang, Yuanhan and Wang, Ziyue and Tan, Haoran and Li, Chunyuan and Liu, Ziwei},
  journal = {arXiv preprint a
             rXiv:2406.16852},
  year    = {2024}
}

@article{chen2024expanding,
  title   = {Expanding performance boundaries of open-source multimodal models with model, data, and test-time scaling},
  author  = {Chen, Zhe and Wang, Weiyun and Cao, Yue and Liu, Yangzhou and Gao, Zhangwei and Cui, Erfei and Zhu, Jinguo and Ye, Shenglong and Tian, Hao and Liu, Zhaoyang and others},
  journal = {arXiv preprint arXiv:2412.05271},
  year    = {2024}
}

@inproceedings{tang2025adaptive,
  title     = {Adaptive keyframe sampling for long video understanding},
  author    = {Tang, Xi and Qiu, Jihao and Xie, Lingxi and Tian, Yunjie and Jiao, Jianbin and Ye, Qixiang},
  booktitle = {CVPR},
  year      = {2025}
}

@inproceedings{park2026too,
  title     = {Too many frames, not all useful: Efficient strategies for long-form video qa},
  author    = {Park, Jongwoo and Ranasinghe, Kanchana and Kahatapitiya, Kumara and Ryu, Wonjeong and Kim, Donghyun and Ryoo, Michael S},
  booktitle = {Proceedings of the 19th Conference of the European Chapter of the Association for Computational Linguistics (Volume 1: Long Papers)},
  pages     = {3569--3588},
  year      = {2026}
}

@article{zhang2025deep,
  title   = {Deep Video Discovery: Agentic Search with Tool Use for Long-form Video Understanding},
  author  = {Zhang, Xiaoyi and Jia, Zhaoyang and Guo, Zongyu and Li, Jiahao and Li, Bin and Li, Houqiang and Lu, Yan},
  journal = {arXiv preprint arXiv:2505.18079},
  year    = {2025}
}

@article{luo2024video,
  title   = {Video-rag: Visually-aligned retrieval-augmented long video comprehension},
  author  = {Luo, Yongdong and Zheng, Xiawu and Li, Guilin and Yin, Shukang and Lin, Haojia and Fu, Chaoyou and Huang, Jinfa and Ji, Jiayi and Chao, Fei and Luo, Jiebo and others},
  journal = {arXiv preprint arXiv:2411.13093},
  year    = {2024}
}

@article{jeong2025videorag,
  title   = {Videorag: Retrieval-augmented generation over video corpus},
  author  = {Jeong, Soyeong and Kim, Kangsan and Baek, Jinheon and Hwang, Sung Ju},
  journal = {arXiv preprint arXiv:2501.05874},
  year    = {2025}
}

@inproceedings{fan2024videoagent,
  title     = {Videoagent: A memory-augmented multimodal agent for video understanding},
  author    = {Fan, Yue and Ma, Xiaojian and Wu, Rujie and Du, Yuntao and Li, Jiaqi and Gao, Zhi and Li, Qing},
  booktitle = {ECCV},
  year      = {2024}
}

@article{zhi2025videoagent2,
  title   = {VideoAgent2: Enhancing the LLM-Based Agent System for Long-Form Video Understanding by Uncertainty-Aware CoT},
  author  = {Zhi, Zhuo and Wu, Qiangqiang and Li, Wenbo and Li, Yinchuan and Shao, Kun and Zhou, Kaiwen and others},
  journal = {arXiv preprint arXiv:2504.04471},
  year    = {2025}
}

@article{liu2024oryx,
  title   = {Oryx mllm: On-demand spatial-temporal understanding at arbitrary resolution},
  author  = {Liu, Zuyan and Dong, Yuhao and Liu, Ziwei and Hu, Winston and Lu, Jiwen and Rao, Yongming},
  journal = {arXiv preprint arXiv:2409.12961},
  year    = {2024}
}

@inproceedings{wang2024videoagent,
  title     = {Videoagent: Long-form video understanding with large language model as agent},
  author    = {Wang, Xiaohan and Zhang, Yuhui and Zohar, Orr and Yeung-Levy, Serena},
  booktitle = {ECCV},
  year      = {2024}
}

@article{yuan2025videodeepresearch,
  title   = {VideoDeepResearch: Long Video Understanding With Agentic Tool Using},
  author  = {Yuan, Huaying and Liu, Zheng and Zhou, Junjie and Wen, Ji-Rong and Dou, Zhicheng},
  journal = {arXiv preprint arXiv:2506.10821},
  year    = {2025}
}

@article{wang2025active,
  title   = {Active Video Perception: Iterative Evidence Seeking for Agentic Long Video Understanding},
  author  = {Wang, Ziyang and Zhou, Honglu and Wang, Shijie and Li, Junnan and Xiong, Caiming and Savarese, Silvio and Bansal, Mohit and Ryoo, Michael S and Niebles, Juan Carlos},
  journal = {arXiv preprint arXiv:2512.05774},
  year    = {2025}
}

@article{kipf2017semi,
  title   = {Semi-supervised classification with graph convolutional networks},
  author  = {Kipf, TN},
  journal = {arXiv preprint arXiv:1609.02907},
  year    = {2016}
}

@article{yedidia2003understanding,
  title   = {Understanding belief propagation and its generalizations},
  author  = {Yedidia, Jonathan S and Freeman, William T and Weiss, Yair and others},
  journal = {Exploring artificial intelligence in the new millennium},
  volume  = {8},
  number  = {236-239},
  year    = {2003}
}

@inproceedings{zhou2003learning,
  title     = {Learning with Local and Global Consistency},
  author    = {Zhou, Dengyong and Bousquet, Olivier and Lal, Thomas Navin and Weston, Jason and Sch{\"o}lkopf, Bernhard},
  booktitle = {NeurIPS},
  volume    = {16},
  year      = {2004}
}

@inproceedings{fu2025video,
  title     = {Video-mme: The first-ever comprehensive evaluation benchmark of multi-modal llms in video analysis},
  author    = {Fu, Chaoyou and Dai, Yuhan and Luo, Yongdong and Li, Lei and Ren, Shuhuai and Zhang, Renrui and Wang, Zihan and Zhou, Chenyu and Shen, Yunhang and Zhang, Mengdan and others},
  booktitle = {CVPR},
  year      = {2025}
}

@inproceedings{wang2025lvbench,
  title     = {Lvbench: An extreme long video understanding benchmark},
  author    = {Wang, Weihan and He, Zehai and Hong, Wenyi and Cheng, Yean and Zhang, Xiaohan and Qi, Ji and Ding, Ming and Gu, Xiaotao and Huang, Shiyu and Xu, Bin and others},
  booktitle = {ICCV},
  year      = {2025}
}

@inproceedings{wu2024longvideobench,
  title     = {Longvideobench: A benchmark for long-context interleaved video-language understanding},
  author    = {Wu, Haoning and Li, Dongxu and Chen, Bei and Li, Junnan},
  booktitle = {NeurIPS},
  year      = {2024}
}

@inproceedings{zhou2025mlvu,
  title     = {Mlvu: Benchmarking multi-task long video understanding},
  author    = {Zhou, Junjie and Shu, Yan and Zhao, Bo and Wu, Boya and Liang, Zhengyang and Xiao, Shitao and Qin, Minghao and Yang, Xi and Xiong, Yongping and Zhang, Bo and others},
  booktitle = {CVPR},
  year      = {2025}
}

@inproceedings{zhai2023sigmoid,
  title     = {Sigmoid loss for language image pre-training},
  author    = {Zhai, Xiaohua and Mustafa, Basil and Kolesnikov, Alexander and Beyer, Lucas},
  booktitle = {ICCV},
  year      = {2023}
}

@inproceedings{li2018deeper,
  title     = {Deeper insights into graph convolutional networks for semi-supervised learning},
  author    = {Li, Qimai and Han, Zhichao and Wu, Xiao-Ming},
  booktitle = {AAAI},
  volume    = {32},
  year      = {2018}
}

@book{chung1997spectral,
  title     = {Spectral Graph Theory},
  author    = {Chung, Fan R. K.},
  year      = {1997},
  publisher = {American Mathematical Society},
  series    = {CBMS Regional Conference Series in Mathematics},
  volume    = {92},
  location  = {Providence, RI},
  doi       = {10.1090/cbms/092},
  isbn      = {978-0-8218-0315-8},
  mrnumber  = {MR1421568}
}

@inproceedings{karpukhin2020dense,
  title     = {Dense Passage Retrieval for Open-Domain Question Answering.},
  author    = {Karpukhin, Vladimir and Oguz, Barlas and Min, Sewon and Lewis, Patrick SH and Wu, Ledell and Edunov, Sergey and Chen, Danqi and Yih, Wen-tau},
  booktitle = {EMNLP},
  year      = {2020}
}

@article{robertson2009probabilistic,
  title   = {The probabilistic relevance framework: BM25 and beyond},
  author  = {Robertson, Stephen and Zaragoza, Hugo and others},
  journal = {Foundations and trends{\textregistered} in information retrieval},
  volume  = {3},
  number  = {4},
  year    = {2009}
}

@article{belkin2006manifold,
  title   = {Manifold regularization: A geometric framework for learning from labeled and unlabeled examples.},
  author  = {Belkin, Mikhail and Niyogi, Partha and Sindhwani, Vikas},
  journal = {Journal of machine learning research},
  volume  = {7},
  number  = {11},
  year    = {2006}
}

@inproceedings{bodla2017soft,
  title     = {Soft-NMS--improving object detection with one line of code},
  author    = {Bodla, Navaneeth and Singh, Bharat and Chellappa, Rama and Davis, Larry S},
  booktitle = {ICCV},
  year      = {2017}
}

@inproceedings{tao2025dycoke,
  title     = {DyCoke: Dynamic Compression of Tokens for Fast Video Large Language Models},
  author    = {Tao, Keda and Qin, Can and You, Haoxuan and Sui, Yang and Wang, Huan},
  booktitle = {CVPR},
  year      = {2025}
}

@article{shen2024longvu,
  title   = {Longvu: Spatiotemporal adaptive compression for long video-language understanding},
  author  = {Shen, Xiaoqian and Xiong, Yunyang and Zhao, Changsheng and Wu, Lemeng and Chen, Jun and Zhu, Chenchen and Liu, Zechun and Xiao, Fanyi and Varadarajan, Balakrishnan and Bordes, Florian and others},
  journal = {arXiv preprint arXiv:2410.17434},
  year    = {2024}
}

@article{hurst2024gpt,
  title   = {Gpt-4o system card},
  author  = {Hurst, Aaron and Lerer, Adam and Goucher, Adam P and Perelman, Adam and Ramesh, Aditya and Clark, Aidan and Ostrow, AJ and Welihinda, Akila and Hayes, Alan and Radford, Alec and others},
  journal = {arXiv preprint arXiv:2410.21276},
  year    = {2024}
}

@article{comanici2025gemini,
  title   = {Gemini 2.5: Pushing the frontier with advanced reasoning, multimodality, long context, and next generation agentic capabilities},
  author  = {Comanici, Gheorghe and Bieber, Eric and Schaekermann, Mike and Pasupat, Ice and Sachdeva, Noveen and Dhillon, Inderjit and Blistein, Marcel and Ram, Ori and Zhang, Dan and Rosen, Evan and others},
  journal = {arXiv preprint arXiv:2507.06261},
  year    = {2025}
}

@article{bai2025qwen2,
  title   = {Qwen2. 5-vl technical report},
  author  = {Bai, Shuai and Chen, Keqin and Liu, Xuejing and Wang, Jialin and Ge, Wenbin and Song, Sibo and Dang, Kai and Wang, Peng and Wang, Shijie and Tang, Jun and others},
  journal = {arXiv preprint arXiv:2502.13923},
  year    = {2025}
}

@inproceedings{arivazhagan2023hybrid,
  title     = {Hybrid hierarchical retrieval for open-domain question answering},
  author    = {Arivazhagan, Manoj Ghuhan and Liu, Lan and Qi, Peng and Chen, Xinchi and Wang, William Yang and Huang, Zhiheng},
  booktitle = {Findings of ACL 2023},
  year      = {2023}
}

@inproceedings{radford2023robust,
  title     = {Robust speech recognition via large-scale weak supervision},
  author    = {Radford, Alec and Kim, Jong Wook and Xu, Tao and Brockman, Greg and McLeavey, Christine and Sutskever, Ilya},
  booktitle = {ICML},
  year      = {2023}
}

@misc{jaidedai_easyocr,
  author       = {{JaidedAI}},
  title        = {EasyOCR},
  year         = {2023},
  howpublished = {\url{https://github.com/JaidedAI/EasyOCR}},
  note         = {Accessed: 2026-01-21}
}

@article{yang2025qwen3,
  title   = {Qwen3 technical report},
  author  = {Yang, An and Li, Anfeng and Yang, Baosong and Zhang, Beichen and Hui, Binyuan and Zheng, Bo and Yu, Bowen and Gao, Chang and Huang, Chengen and Lv, Chenxu and others},
  journal = {arXiv preprint arXiv:2505.09388},
  year    = {2025}
}

@article{bai2025qwen3vltechnicalreport,
  title   = {{Q}wen3-{VL} {T}echnical {R}eport},
  author  = {Shuai Bai and Yuxuan Cai and Ruizhe Chen and Keqin Chen and Xionghui Chen and others},
  year    = {2025},
  journal = {arXiv preprint arXiv:2511.21631}
}

@article{guo2025seed1,
  title   = {Seed1. 5-vl technical report},
  author  = {Guo, Dong and Wu, Faming and Zhu, Feida and Leng, Fuxing and Shi, Guang and Chen, Haobin and Fan, Haoqi and Wang, Jian and Jiang, Jianyu and Wang, Jiawei and others},
  journal = {arXiv preprint arXiv:2505.07062},
  year    = {2025}
}

@article{team2024gemini,
  title   = {Gemini 1.5: Unlocking multimodal understanding across millions of tokens of context},
  author  = {Team, Gemini and Georgiev, Petko and Lei, Ving Ian and Burnell, Ryan and Bai, Libin and Gulati, Anmol and Tanzer, Garrett and Vincent, Damien and Pan, Zhufeng and Wang, Shibo and others},
  journal = {arXiv preprint arXiv:2403.05530},
  year    = {2024}
}

@article{zhang2024video,
  title   = {Video instruction tuning with synthetic data},
  author  = {Zhang, Yuanhan and Wu, Jinming and Li, Wei and Li, Bo and Ma, Zejun and Liu, Ziwei and Li, Chunyuan},
  journal = {arXiv preprint arXiv:2410.02713},
  year    = {2024}
}

@inproceedings{wang2025adaretake,
  title     = {AdaReTaKe: Adaptive redundancy reduction to perceive longer for video-language understanding},
  author    = {Wang, Xiao and Si, Qingyi and Zhu, Shiyu and Wu, Jianlong and Cao, Li and Nie, Liqiang},
  booktitle = {Findings of ACL 2025},
  year      = {2025}
}

@inproceedings{radford2021learning,
  title        = {Learning transferable visual models from natural language supervision},
  author       = {Radford, Alec and Kim, Jong Wook and Hallacy, Chris and Ramesh, Aditya and Goh, Gabriel and Agarwal, Sandhini and Sastry, Girish and Askell, Amanda and Mishkin, Pamela and Clark, Jack and others},
  booktitle    = {ICML},
  year         = {2021},
  organization = {PmLR}
}

@article{achiam2023gpt,
  title   = {Gpt-4 technical report},
  author  = {Achiam, Josh and Adler, Steven and Agarwal, Sandhini and Ahmad, Lama and Akkaya, Ilge and Aleman, Florencia Leoni and Almeida, Diogo and Altenschmidt, Janko and Altman, Sam and Anadkat, Shyamal and others},
  journal = {arXiv preprint arXiv:2303.08774},
  year    = {2023}
}

@article{belkin2003laplacian,
  title   = {Laplacian eigenmaps for dimensionality reduction and data representation},
  author  = {Belkin, Mikhail and Niyogi, Partha},
  journal = {Neural computation},
  volume  = {15},
  number  = {6},
  year    = {2003}
}

@article{liu2024deepseek,
  title   = {Deepseek-v3 technical report},
  author  = {Liu, Aixin and Feng, Bei and Xue, Bing and Wang, Bingxuan and Wu, Bochao and Lu, Chengda and Zhao, Chenggang and Deng, Chengqi and Zhang, Chenyu and Ruan, Chong and others},
  journal = {arXiv preprint arXiv:2412.19437},
  year    = {2024}
}

@inproceedings{lin2024video,
  title     = {Video-llava: Learning united visual representation by alignment before projection},
  author    = {Lin, Bin and Ye, Yang and Zhu, Bin and Cui, Jiaxi and Ning, Munan and Jin, Peng and Yuan, Li},
  booktitle = {EMNLP},
  year      = {2024}
}

@article{fu2025vita,
  title   = {Vita-1.5: Towards gpt-4o level real-time vision and speech interaction},
  author  = {Fu, Chaoyou and Lin, Haojia and Wang, Xiong and Zhang, Yi-Fan and Shen, Yunhang and Liu, Xiaoyu and Cao, Haoyu and Long, Zuwei and Gao, Heting and Li, Ke and others},
  journal = {arXiv preprint arXiv:2501.01957},
  year    = {2025}
}

@article{li2024llava,
  title   = {Llava-onevision: Easy visual task transfer},
  author  = {Li, Bo and Zhang, Yuanhan and Guo, Dong and Zhang, Renrui and Li, Feng and Zhang, Hao and Zhang, Kaichen and Zhang, Peiyuan and Li, Yanwei and Liu, Ziwei and others},
  journal = {arXiv preprint arXiv:2408.03326},
  year    = {2024}
}

@article{chen2025scaling,
  title   = {Scaling rl to long videos},
  author  = {Chen, Yukang and Huang, Wei and Shi, Baifeng and Hu, Qinghao and Ye, Hanrong and Zhu, Ligeng and Liu, Zhijian and Molchanov, Pavlo and Kautz, Jan and Qi, Xiaojuan and others},
  journal = {arXiv preprint arXiv:2507.07966},
  year    = {2025}
}

@article{wang2025models,
  title     = {From Models to Systems: A Comprehensive Survey of Efficient Multimodal Learning},
  author    = {Wang, Pan and Song, Siwei and Ji, Hui and Cao, Siqi and Yu, Heng and Liu, Zhijian and others},
  journal   = {Authorea Preprints},
  year      = {2025},
  publisher = {Authorea}
}

@article{liu2025towards,
  title   = {Towards training-free long video understanding: methods, benchmarks, and open challenges},
  author  = {Liu, Jingren and Wang, Yun and Zhang, Long and others},
  year    = {2025},
  journal = {Vicinagearth},
  volume  = {2},
  number  = {6},
  doi     = {10.1007/s44336-025-00017-w}
}


\clearpage
\appendix

\section{Belief Propagation: Theoretical Analysis}
\label{app:diffusion}

\subsection{Closed-form Solution}
The iterative diffusion process in Eq.~\eqref{eq:diff_iter} converges to a closed-form solution. After infinite iterations, the belief field converges to:
$\vecv{F}^\star=(1-\beta)\left(\mat{I}-\beta\,\mat{W}_{\mathrm{norm}}\right)^{-1}\vecv{Y},$
where $\mat{I}$ is the identity matrix. This can be derived by setting $\vecv{F}^{(t+1)} = \vecv{F}^{(t)} = \vecv{F}^\star$ and solving for $\vecv{F}^\star$.

\subsection{Convergence Analysis}

The spectral radius of the symmetric normalized affinity matrix $\mat{W}_{\mathrm{norm}}$ is bounded by 1 due to the normalization in Eq.~\eqref{eq:wnorm}. This ensures that the iterative process converges exponentially fast. Specifically, let $\lambda_{\max}$ denote the largest eigenvalue of $\mat{W}_{\mathrm{norm}}$. The convergence rate is determined by $\beta\lambda_{\max} < 1$, which guarantees stability.

\subsection{Computational Efficiency}

Direct matrix inversion to obtain the closed-form solution requires $O(K^3)$ operations. In contrast, with top-$k$ sparsification and sparse matrix-vector multiplication, the iterative approach requires $O(TKk)$ operations, where $T$ is the number of iterations (typically $T \ll K$ and $k \ll K$). If implemented with dense matrix operations, the cost becomes the looser $O(TK^2)$ upper bound. More importantly, when a new observation arrives and updates $\vecv{Y}^{(t)}$, we can continue iterating from the current state $\vecv{F}^{(t)}$ without recomputing from scratch, enabling efficient incremental updates crucial for active learning.

\section{Evidence Selection: Detailed Algorithm}
\label{app:selection}

\subsection{Graph-NMS Algorithm}

To avoid selecting redundant evidence from spatially-temporally adjacent segments, we employ a Graph-NMS procedure that suppresses neighbors of already-selected nodes (Alg.~\ref{alg:graphnms}).

\begin{algorithm}[t]
\caption{Graph-NMS for Evidence Selection}
\label{alg:graphnms}
\begin{algorithmic}[1]
\REQUIRE Final belief field $\vecv{F}^{(T)}$, prior channels $\{Y^{\mathrm{prior}}_r\}_{r=1}^R$, affinity matrix $\tilde{\mat{W}}$, suppression factor $\eta\in(0,1)$, number of nodes to select $m$
\ENSURE Selected node set $\mathcal{S}$
\STATE Initialize $\mathcal{S} \leftarrow \emptyset$, $\vecv{F}' \leftarrow \vecv{F}^{(T)}$
\STATE \textbf{// Ensure each facet has at least one representative}
\FOR{$r=1$ to $R$}
    \STATE $i_r \leftarrow \argmax_i (Y^{\mathrm{prior}}_r)_i \cdot F'^{}_i$
    \STATE $\mathcal{S} \leftarrow \mathcal{S} \cup \{i_r\}$
\ENDFOR
\STATE \textbf{// Iteratively select high-confidence nodes}
\WHILE{$|\mathcal{S}| < m$}
    \STATE $i^\star \leftarrow \argmax_{i\notin\mathcal{S}} F'_i$
    \IF{$F'_{i^\star} \le 0$}
        \STATE \textbf{break} \COMMENT{No more positive confidence nodes}
    \ENDIF
    \STATE $\mathcal{S} \leftarrow \mathcal{S} \cup \{i^\star\}$
    \STATE \textbf{// Suppress neighbors}
    \FOR{each neighbor $j \in \mathcal{N}(i^\star)$ with $\tilde{W}_{i^\star j} > 0$}
        \STATE $F'_j \leftarrow \eta \cdot F'_j$
    \ENDFOR
\ENDWHILE
\STATE \textbf{return} $\mathcal{S}$
\end{algorithmic}
\end{algorithm}

The suppression factor $\eta$ controls the strength of neighbor suppression. A smaller $\eta$ leads to more aggressive suppression, encouraging selection of nodes that are more dispersed in the graph. In our experiments, we set $\eta = 0.2$.

\subsection{Evidence Packaging Details}

For each selected node $i \in \mathcal{S}$, we construct a compact multimodal evidence package consisting of:

\begin{itemize}
\item \textbf{Visual frames}: Sample $n_f$ representative frames uniformly from the time span $[s_i, e_i]$. In practice, we use $n_f = 4$ for computational efficiency.
\item \textbf{Best textual evidence}: Among the three evidence sources (caption $e_i^{\mathrm{cap}}$, OCR text $e_i^{\mathrm{ocr}}$, ASR text $e_i^{\mathrm{asr}}$), select the one with the highest relevance score computed in \S\ref{sec:scoring}:
\begin{equation}
e_i^{\mathrm{best}} = \argmax_{e \in \{e_i^{\mathrm{cap}}, e_i^{\mathrm{ocr}}, e_i^{\mathrm{asr}}\}} \max_{r} s(e, f_r).
\end{equation}
This ensures we include only the most relevant textual evidence while avoiding redundancy.
\item \textbf{Temporal information}: The start and end timestamps $[s_i, e_i]$ to maintain temporal ordering.
\end{itemize}

These packages are sorted by temporal order and concatenated into a structured prompt for the downstream MLLM, which generates the final answer based on the aggregated evidence.

\section{Prompts for LLM and VLM Calls}
\label{app:prompts}

This section provides the core prompts used in our implementation.

\subsection{Query Decomposition Prompt (LLM)}

The LLM decomposes the query into entities (for keyword matching) and events (for semantic matching).

\begin{tcolorbox}[promptbox, title={\small\textbf{System Prompt}}]
\small
\textbf{Role:} Entity \& Event Extractor for Video Understanding.\\
\textbf{Task:} Extract ENTITIES (for keyword matching) and EVENTS (for semantic matching) from user query.\\[3pt]
\textbf{Output Requirements:}\\
1. \textbf{Query Keywords (ENTITIES):} person names, place names, object names, numbers, years.\\
2. \textbf{General Semantic Query (EVENT):} what event must happen to answer the question.\\
3. \textbf{Option Keywords:} 2-5 specific entities per option.\\
4. \textbf{Option Semantic Queries:} the specific event indicating each option is correct.\\[3pt]
\textbf{Rules:} Keywords = ENTITIES; Semantic Queries = EVENTS. At least one must be non-empty per option.
\end{tcolorbox}

\begin{tcolorbox}[promptbox, title={\small\textbf{User Prompt}}]
\small
Input Query: ``\{query\}''\\
Extract ENTITIES and EVENTS for video retrieval.\\[3pt]
\textbf{Output JSON format:}
\begin{lstlisting}
{
  "query_keywords": [...],
  "option_keywords": {"A": [...], ...},
  "semantic_queries": {"A": "...", ...},
  "general_semantic_query": "...",
  "temporal_plan": "...",
  "vlm_query": "..."
}
\end{lstlisting}
\end{tcolorbox}

\subsection{Observer Inspection Prompt (VLM)}

The VLM observes a video segment and generates a caption plus logical gap analysis.

\begin{tcolorbox}[promptbox, title={\small\textbf{System Prompt}}]
\small
\textbf{Role:} Video Captioner \& Logic Analyst.\\
\textbf{Task:} Analyze the clip and output a comprehensive caption plus missing visual evidence.\\[3pt]
\textbf{Your job:}\\
1.\textbf{Caption:} Describe what is visible comprehensively (WHO/WHAT/WHERE, actions, objects, text).\\
2.\textbf{Logical Gap:} If the answer is not here, what specific visual event is missing?\\[3pt]
\textbf{Rules:} Quote on-screen text accurately if visible.\\[3pt]
\textbf{Output JSON format:}
\begin{lstlisting}
{
  "reasoning": "...",
  "caption": "...",
  "refinement_plan": {
    "needs_more_info": bool,
    "missing_visual_keyword": "..."
  }
}
\end{lstlisting}
\end{tcolorbox}

\begin{tcolorbox}[promptbox, title={\small\textbf{User Prompt}}]
\small
Query: \{query\}\\
Focus Keywords: \{focus\_keywords\}\\
Focus Semantic Queries: \{focus\_semantic\_queries\}\\[3pt]
Analyze these video frames and determine: (1) What is visible in this clip; (2) If we need more info, what specific visual should we look for?\\
Output only valid JSON.
\end{tcolorbox}

\subsection{Final Answer Generation Prompt (VLM) When Evaluating}

The VLM generates the final answer based on selected evidence frames.

\begin{tcolorbox}[promptbox, title={\small\textbf{System Prompt}}]
\small
You are a video analysis assistant. Identify the ONE \{criteria\} statement among the options.\\[3pt]
\textbf{Decision Steps:}\\
1. Read the question and all options.\\
2. For each option, check frames and attached evidence lines.\\
3. Prefer explicit evidence over vague impressions.\\
4. For order/time questions, compare early vs late frames; for text, use OCR evidence.\\
5. If evidence is weak, choose most plausible option and state low confidence.\\[3pt]
\textbf{Rules:} MUST output an option LETTER (A/B/C/D). DO NOT output ``NO EVIDENCE''.\\[3pt]
\textbf{Response Format:}\\
Analysis: \textless your reasoning\textgreater\\
Final Answer: \textless ONE LETTER\textgreater\\
Reason: \textless one short sentence\textgreater
\end{tcolorbox}

\begin{tcolorbox}[promptbox, title={\small\textbf{User Prompt}}]
\small
Based on these video frames, answer the following question:\\[3pt]
Frame Information: \{frame\_info\_str\}\\
Question: \{query\}\\[3pt]
Remember: include a clear ``Final Answer: \textless LETTER\textgreater'' line so it can be parsed.
\end{tcolorbox}

\section{Implementation Details and Hyperparameters}
\label{app:hyperparams}

This section provides complete hyperparameter settings used in our experiments. All parameters are consistent across all benchmarks unless otherwise specified.

\subsection{Backbone Comparison Experimental Configuration}
\label{app:backbone_config}

For the backbone comparison experiments shown in Figure~\ref{fig:backbone_comparison}, we use the following configurations:

\textbf{Frame Sampling}:
\begin{itemize}
\item VideoXL2, Oryx-1.5: 16 frames
\item InternVL-2.5: 8 frames  
\item All other models: 32 frames
\end{itemize}

\textbf{LLM configuration}:
\begin{itemize}
\item GLM, SeedVL, and Qwen3-VL (30B/32B variants): Qwen3-30B as the LLM planner
\item All other models: Qwen3-8B as the LLM planner
\end{itemize}

These configurations ensure that each model is tested under its optimal or commonly used settings while maintaining fairness in comparison. The varying frame sampling reflect the different input capacity and design of each model, and the LLM selection is matched to the scale of the visual backbone for computational efficiency.

\subsection{Main Results Table Configuration}
\label{app:main_results_config}

For the main comparison results shown in Table~\ref{tab:main_results}, we instantiate VideoDetective with two configurations to demonstrate its effectiveness across different parameter scales:

\textbf{Lightweight Setting ($<$30B):}
\begin{itemize}
\item Visual-Language Model (VLM): Qwen3-VL-8B-Instruct
\item Language Model (LLM) Planner: Qwen3-8B-Instruct
\item Final answer frame sampling: 32 frames
\end{itemize}

\textbf{Larger-scale Setting ($\ge$30B):}
\begin{itemize}
\item Visual-Language Model (VLM): SeedVL-1.5
\item Language Model (LLM) Planner: Qwen3-30B-Instruct
\item Final answer frame sampling: 32 frames
\end{itemize}

Both configurations share the same hyperparameters for graph construction, belief propagation, and active inference as specified in subsequent sections. The frame budget for final answer generation is fixed at 32 frames across both settings to ensure fair comparison with baseline models.

\subsection{Token Efficiency Data Collection}
\label{app:token_efficiency}

For the token efficiency analysis shown in Figure~\ref{fig:efficiency}, we report the average token consumption per video on VideoMME-long. The data is collected through the following methods:

\textbf{Experimental Measurements (Method Baselines):}
\begin{itemize}
\item \textbf{VideoAgent, DVD, LVNet, and VideoDetective}: The token counts are directly obtained from real experimental runs via API response data. These values represent the actual token consumption during inference.
\end{itemize}

\textbf{Estimated Lower Bounds (Model Baselines):}
\begin{itemize}
\item \textbf{Gemini-1.5-Pro, GPT-4o, and LLaVA-Video-72B}: We estimate the \textit{lower bound} of token consumption based on:
  \begin{enumerate}
  \item Official sampling rates (frames per video)
  \item Per-frame token counts specified in official API documentation
  \item Standard video resolution settings
  \end{enumerate}
\end{itemize}

\textbf{Important Notes:}
\begin{itemize}
\item These estimates include \textit{only image tokens} and exclude text prompts, system instructions, and other textual overhead.
\item This makes them conservative baselines---the actual token consumption of these models would be higher in practice.
\item All measurements are averaged across all videos in the VideoMME-long benchmark.
\end{itemize}

\subsection{Lexical and Semantic Similarity Computation}
\label{app:similarity}

We provide detailed implementation of the lexical and semantic similarity scores used in evidence scoring (\S\ref{sec:scoring}).

\subsubsection{Motivation: Complementary Sparse-Dense Retrieval}
\label{app:sparse_dense}

Sparse retrieval (lexical matching) and dense retrieval (semantic matching) provide \emph{complementary inductive biases}~\cite{robertson2009probabilistic,karpukhin2020dense}:

\begin{itemize}
\item \textbf{Dense vectors (embeddings)}: Excel at handling synonym paraphrasing and semantic equivalence (e.g., "automobile" $\approx$ "car"), enabling robust generalization. However, they are susceptible to ``semantic drift''—embeddings may conflate related but distinct concepts, leading to false positives (high recall, lower precision).

\item \textbf{Sparse lexical matching (IDF-weighted overlap)}: Ensure symbol-level precision by exact token matching (e.g., distinguishing "bank" as financial institution vs. riverbank). However, it is insensitive to paraphrasing and synonyms (high precision, low recall). This score is TF-IDF-inspired but does not require constructing full TF-IDF vectors.
\end{itemize}

By combining both approaches with source-aware weighting (\S\ref{sec:lexsem}), we achieve both precision and recall: lexical matching captures exact mentions while semantic matching handles variations and implicit references.

\subsubsection{IDF-weighted Lexical Overlap (Sparse Matching)}

For lexical matching, we use an IDF-weighted lexical overlap score with standard preprocessing:
\begin{enumerate}
\item \textbf{Text preprocessing}: Lowercase conversion, stopword removal, and lemmatization.
\item \textbf{IDF computation}: Pre-computed on a large corpus, with out-of-vocabulary words assigned a default IDF value.
\item \textbf{Score computation}: For evidence text $e$ and keyword set $\mathcal{K}_r$:
\[
s_{\mathrm{lex}}(e, f_r) = \min\left(1.0, \frac{\sum_{t \in e \cap \mathcal{K}_r} \text{IDF}(t)}{Z_{\mathrm{lex}}}\right)
\]
where $Z_{\mathrm{lex}}=3.0$ is a normalization constant.
\item \textbf{Normalization}: We clip scores to $[0,1]$ via the $\min(\cdot)$ term above.
\end{enumerate}

\subsubsection{Embedding-based Semantic Similarity}

For semantic matching, we use SigLIP text encoder with cosine similarity:
\begin{enumerate}
\item \textbf{Text encoding}: $\psi(e) = \text{SigLIP-Text}(e) \in \mathbb{R}^d$, $\|\psi(e)\|_2 = 1$.
\item \textbf{Score computation}: For evidence text $e$ and semantic query set $\mathcal{P}_r$, we compute:
\[
s_{\mathrm{sem}}(e, f_r) = \max_{p \in \mathcal{P}_r} \langle \psi(e), \psi(p) \rangle
\]
where $p$ represents semantic queries (event descriptions) that capture the contextual meaning of each facet.
\item \textbf{Batch encoding}: All semantic queries are pre-encoded for efficiency.
\end{enumerate}

\subsection{Source-aware Fusion}

Different evidence sources have different signal-to-noise characteristics:
\begin{itemize}
\item \textbf{OCR text}: High precision, low recall. Weight: $\lambda_{\mathrm{ocr}} = 0.7$ (trust lexical more).
\[
s_{\mathrm{ocr}}(e, f_r) = 0.7 \cdot s_{\mathrm{lex}}(e, f_r) + 0.3 \cdot s_{\mathrm{sem}}(e, f_r)
\]
\item \textbf{ASR text}: Balanced. Weight: $\lambda_{\mathrm{asr}} = 0.5$ (equal trust).
\[
s_{\mathrm{asr}}(e, f_r) = 0.5 \cdot s_{\mathrm{lex}}(e, f_r) + 0.5 \cdot s_{\mathrm{sem}}(e, f_r)
\]
\item \textbf{Caption}: High recall, may generalize. Weight: $\lambda_{\mathrm{cap}} = 0.3$ (trust semantic more).
\[
s_{\mathrm{cap}}(e, f_r) = 0.3 \cdot s_{\mathrm{lex}}(e, f_r) + 0.7 \cdot s_{\mathrm{sem}}(e, f_r)
\]
\end{itemize}

Final node score: $s_i = \max_{e\in E_i,\, r} s(e, f_r)$.

\subsubsection{Event Description Generation for Semantic Channel}
\label{app:event_skeleton}

This section details how the event descriptions $\{e_i\}$ are generated, which are used in the \textbf{Hypothesis stage} (\S\ref{sec:inference}) for multi-route prior initialization. Specifically, in Eq.~\eqref{eq:prior_init}, the semantic query $p \in \mathcal{P}_r$ is matched against these event descriptions to compute the semantic channel of the prior score.

\textbf{Generation process}:
\begin{enumerate}
\item \textbf{Uniform sampling}: Extract $F$ frames uniformly distributed across the \emph{entire video} (not per-node). We reuse the same frame sampling number $F$ as the final answer generation.
\item \textbf{VLM generation (time-stamped event timeline)}: Use the VLM to generate a coarse event timeline based on these $F$ frames, capturing the overall narrative and key events. Concretely, the VLM outputs a list of event items, each with an approximate temporal span (e.g., start/end timestamps or the corresponding frame indices among the $F$ sampled frames) plus a short textual description.
\item \textbf{Deterministic node-level assignment}: Each node corresponds to a video chunk with a temporal interval $[s_i, e_i]$. We assign to node $i$ all event items whose temporal spans overlap with $[s_i, e_i]$ (or whose associated sampled-frame indices fall within the node's interval), and concatenate their descriptions to form $e_i$. If no event item overlaps, we assign the temporally nearest event item (by midpoint distance) as $e_i$.
\end{enumerate}

\textbf{Important notes}:
\begin{itemize}
\item This event description is \textbf{coarse-grained} and serves as a \textbf{semantic complement} to the keyword-based (cross-modal) channel in the multi-route prior.
\item It helps capture high-level event semantics that pure keyword matching may miss (e.g., ``A person explains X before demonstrating Y'').
\item In practice, we set \texttt{skeleton\_frames=$F$} and use the same VLM backbone for consistency.
\end{itemize}

\subsection{Graph Construction and Propagation}

\begin{table}[H]
\centering
\caption{Graph construction and belief propagation parameters.}
\small
\begin{tabular}{lcc}
\toprule
\textbf{Parameter} & \textbf{Symbol} & \textbf{Value} \\
\midrule
Visual-temporal fusion weight & $\alpha$ & 0.6 \\
Temporal decay factor & $\tau$ & 30.0 \\
Top-$k$ sparsification & $k$ & 8 \\
Scene boundary threshold & $\theta_{\mathrm{sim}}$ & 0.82 \\
Minimum chunk length & $L_{\min}$ & 10 frames \\
Propagation iterations & $T_{\mathrm{prop}}$ & 7 \\
Diffusion smoothness parameter & $\beta$ & 0.6 \\
\bottomrule
\end{tabular}
\end{table}
\subsection{Active Inference and Observation}

\begin{table}[H]
\centering
\caption{Active inference and observation parameters.}
\small
\begin{tabular}{lcc}
\toprule
\textbf{Parameter} & \textbf{Symbol} & \textbf{Value} \\
\midrule
Final answer frame sampling & $F$ & 32 frames \\
Base max steps & -- & 10 \\
Steps per extra option & -- & 1 \\
Local observation window & -- & 9 frames \\
Retry relevance threshold & -- & 0.2 \\
Fallback max relevance threshold & -- & 0.4 \\
Fallback mean relevance threshold & -- & 0.2 \\
Flat gap threshold & -- & 0.15 \\
Multi-route fusion weight & $\alpha_{\mathrm{route}}$ & 0.5 \\
\bottomrule
\end{tabular}
\end{table}
\subsection{Evidence Selection and Scoring}

\begin{table}[H]
\centering
\caption{Evidence selection and scoring parameters.}
\scriptsize
\setlength{\tabcolsep}{2.6pt}
\renewcommand{\arraystretch}{1.05}
\begin{tabular}{@{}p{0.38\linewidth} p{0.16\linewidth} p{0.45\linewidth}@{}}
\toprule
\textbf{Parameter} & \textbf{Symbol} & \textbf{Value} \\
\midrule
Number of chunks to select & $m$ & 8 \\
Frames per chunk & $n_f$ & 4 \\
Minimum uniform frames & -- & 4 \\
Graph-NMS suppression factor & $\eta$ & 0.2 \\
Frame deduplication threshold & -- & 0.92 \\
Relaxed deduplication threshold & -- & 0.95 \\
Fallback similarity threshold & -- & 0.90 \\
\midrule
\multicolumn{3}{l}{\textit{Source-aware fusion weights}} \\
OCR text weight & $\lambda_{\mathrm{ocr}}$ & 0.7 (lex) + 0.3 (sem) \\
ASR text weight & $\lambda_{\mathrm{asr}}$ & 0.5 (lex) + 0.5 (sem) \\
Caption weight & $\lambda_{\mathrm{cap}}$ & 0.3 (lex) + 0.7 (sem) \\
Lexical normalization constant & $Z_{\mathrm{lex}}$ & 3.0 (clip to [0,1]) \\
\bottomrule
\end{tabular}
\end{table}
\subsection{Model Configuration}

\textbf{Multi-source evidence generation}: During observation of node $i$, we extract three complementary evidence sources: (1) \textbf{VLM caption}: the VLM generates a textual description of the visual content in sampled frames; (2) \textbf{OCR text}: EasyOCR extracts any on-screen text visible in the frames; (3) \textbf{ASR transcript}: Whisper provides pre-generated speech transcripts for the corresponding time segment. These three sources are scored independently via lexical-semantic matching (\S\ref{sec:lexsem}), and the maximum score is used as the node's relevance: $s_i = \max\{s_{\mathrm{ocr}}, s_{\mathrm{asr}}, s_{\mathrm{cap}}\}$.

\begin{table}[H]
\centering
\caption{Model configurations for main experiments (Qwen3-30B + SeedVL-1.5).}
\small
\setlength{\tabcolsep}{4pt}
\begin{tabular}{lc}
\toprule
\textbf{Component} & \textbf{Configuration} \\
\midrule
\multicolumn{2}{l}{\textit{Visual-Language Model (VLM)}} \\
Model & SeedVL-1.5 \\
Max tokens & 4096 \\
Temperature & 0.0 \\
Timeout & 300s \\
\midrule
\multicolumn{2}{l}{\textit{Text Language Model (LLM)}} \\
Model & Qwen3-30B-Instruct \\
Max tokens & 2048 \\
Temperature & 0.0 \\
\midrule
\multicolumn{2}{l}{\textit{Visual Encoder}} \\
Image encoder & SigLIP-SO400M-patch14-384 \\
Text encoder & SigLIP (text tower) \\
Max text length & 64 tokens \\
\midrule
\multicolumn{2}{l}{\textit{Evidence Extraction Tools}} \\
VLM caption & SeedVL-1.5 (visual description) \\
OCR extraction & EasyOCR (on-screen text) \\
ASR transcription & Whisper (speech-to-text) \\
\midrule
\multicolumn{2}{l}{\textit{Preprocessing}} \\
Sampling rate & 1.0 FPS \\
Cache enabled & Yes \\
\bottomrule
\end{tabular}
\end{table}

\subsection{Retry and Error Handling}

\begin{table}[H]
\centering
\caption{Retry mechanism parameters for API calls.}
\small
\begin{tabular}{lc}
\toprule
\textbf{Parameter} & \textbf{Value} \\
\midrule
Max retry attempts & 5 \\
Base retry delay & 1.0s \\
Max retry delay & 20.0s \\
\bottomrule
\end{tabular}
\end{table}

\end{document}